\documentclass[10pt,twocolumn,letterpaper]{article}

\usepackage{cvpr}              % To produce the CAMERA-READY version
\usepackage{tcolorbox} 
\usepackage{booktabs}       % 用于 \toprule, \midrule
\usepackage{multirow}       % 用于 \multirow
\usepackage{graphicx}       % 用于 \resizebox
\usepackage[table]{xcolor}  % 用于 \cellcolor，一定要加 [table]
\usepackage{amssymb}        % 用于 \checkmark (勾选符号)

\definecolor{cvprblue}{rgb}{0.21,0.49,0.74}
\usepackage[pagebackref,breaklinks,colorlinks,allcolors=cvprblue]{hyperref}
\usepackage{multirow}

\usepackage[table]{xcolor}
\definecolor{oai-gray-600}{RGB}{140,140,140}
\definecolor{oai-gray-300}{RGB}{210,210,210}
\definecolor{oai-green-200}{RGB}{180,230,180}
\definecolor{oai-green-400}{RGB}{140,210,140}
\definecolor{oai-green-600}{RGB}{100,190,100}
\definecolor{navyblue}{RGB}{30,60,120}
\def\paperID{} % *** Enter the Paper ID here
\def\confName{CVPR}
\def\confYear{2026}

\title{Map2Thought: Explicit 3D Spatial Reasoning via Metric Cognitive Maps}

\vspace{-5pt}
\author{
Xiangjun Gao$^1$ 
\quad 
Zhensong Zhang$^{2}$ 
\quad 
Dave Zhenyu Chen$^2$
\quad 
Songcen Xu$^2$ 
\quad 
\\
Long Quan$^1$
\quad 
Eduardo P\'erez-Pellitero$^2$
\quad 
Youngkyoon Jang$^2$ \\
\vspace{-3pt}\normalsize $^1$The Hong Kong University of Science and Technology
\quad 
$^2$Huawei Noah’s Ark Lab
\vspace{10pt}
\\
}

\begin{document}

\twocolumn[\maketitle\vspace{-4em}%#################################################################################################

\begin{center}
\includegraphics[width=0.965\linewidth]{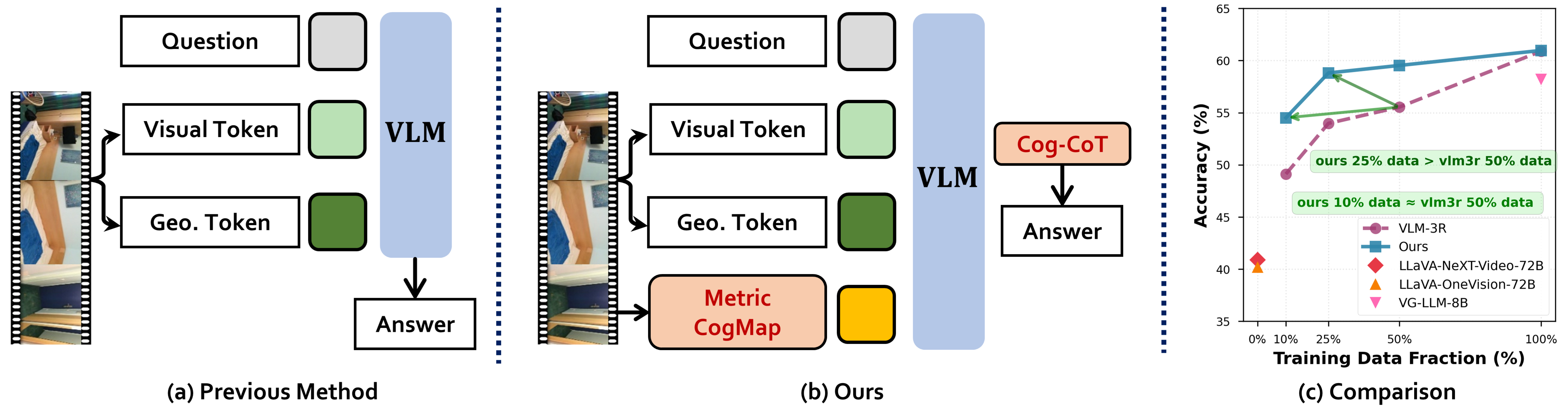}
\end{center} \vspace{-1.5em}
\captionof{figure}{
\textbf{Comparison with prior approaches and efficiency gains.} (a) Previous 3D VLMs fuse visual and geometric tokens but through implicit latent reasoning, limiting spatial interpretability. (b) Our approach introduces a metrically grounded cognitive map (\emph{Metric-CogMap}) and a chain-of-thought-style reasoning process (\emph{Cog-CoT}), enabling explicit and interpretable spatial reasoning from the same multimodal inputs. (c) This design yields substantial data efficiency: with just 10\% or 25\% of the training data, our model is comparable to or surpasses the performance of the baseline model trained with 50\% of the data.
}
\label{fig:teaser}

\bigbreak]

\begin{abstract}

We propose Map2Thought, a framework that enables explicit and interpretable spatial reasoning for 3D VLMs. The framework is grounded in two key components: Metric Cognitive Map ({Metric-CogMap}) and Cognitive Chain-of-Thought ({Cog-CoT}). {Metric-CogMap} provides a unified spatial representation by integrating a discrete grid for relational reasoning with a continuous, metric-scale representation for precise geometric understanding. Building upon the {Metric-CogMap}, {Cog-CoT} performs explicit geometric reasoning through deterministic operations (e.g., vector operations, bounding-box distances, and occlusion-aware appearance order cues)
%—such as vector differences, dot and cross products, bounding-box distances, and occlusion-aware appearance order cues—
producing interpretable inference traces grounded in 3D structure.
%\dave{this reads technically weak - is there any more technical details to strengthen this sentence?}
Experimental results show that Map2Thought enables explainable 3D understanding, achieving $59.9\%$ accuracy using only half the supervision—closely matching the $60.9\%$ baseline trained with full dataset. It consistently outperforms state-of-the-art methods by $5.3\%$, $4.8\%$, and $4.0\%$ under $10\%$, $25\%$, and $50\%$ training subsets, respectively, on the VSI-Bench.
%\dave{please quantify the performance, e.g. our method outperforms the previous method by xxx\% using only yyy\% training data}

\end{abstract}

% \begingroup
% \renewcommand\thefootnote{}
% \footnotetext{Work done during an internship at Huawei Noah’s Ark Lab, London, UK.}
% \endgroup

\section{Introduction}
\label{sec:intro}

Recent advances in 3D Vision-Language Models (3D-VLMs) have increasingly explored bridging multi-modal input signals to harness the rich knowledge priors of large language models (LLMs) for 3D spatial understanding. Beyond projecting 3D geometric tokens from point clouds~\cite{leo, Inst3D-LMM} or RGB-D sequences~\cite{llava3d, video3dllm} into the latent space of large language models, recent advances have leveraged pretrained 3D vision foundation models—such as VGGT~\cite{wang2025vggt} and CUT3R~\cite{cut3r}—which extract 3D structure directly from monocular RGB videos to support more grounded and scalable spatial reasoning. This integration allows 3D-VLMs to access geometry-aware representations without relying on additional modalities. While these methods represent a step forward in generalizability compared to earlier spatial reasoning techniques~\cite{azuma2022scanqa,ye20223d, chen2020scanrefer, chen2021scan2cap, chen2022d3net, chen2023unit3d, dwedari2023generating}, several core challenges remain unresolved:
\begin{itemize}
\item \textbf{Implicit fusion without geometric grounding:} Multimodal representations are fused implicitly without enforcing alignment with physical constraints, limiting transparency and verifiability in spatial reasoning.

\item \textbf{Lack of accumulated spatial context:} Temporal and scene-level spatial cues remain weakly supervised, limiting the model’s ability to reason about appearance order and global layout.

\item \textbf{Biased supervision and poor generalization:} Training on subsampled or biased visual inputs leads to overfitting, impairing reasoning about object scale, type, and spatial anchoring across diverse scene types.
\end{itemize}

\begin{figure*}[!t]
	\centering
	\includegraphics[width=1.0\linewidth]{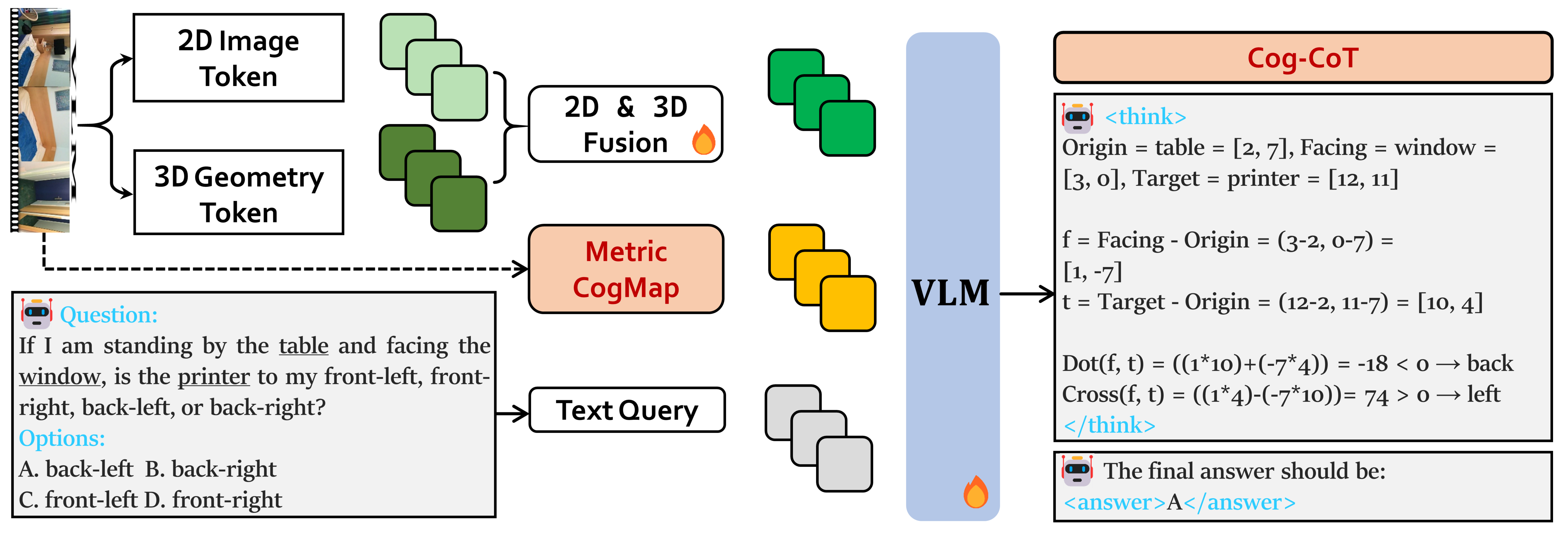}
        \caption{\textbf{Overview of our method.} Given an RGB video and a language query, we extract 2D image tokens and 3D geometry-aware tokens, fuse them into 3D-aware visual tokens, and input them to the VLM. The Metric-CogMap (orange blocks) encodes the scene using both a discrete grid and a metric-scale spatial representation. Cog-CoT (right, grey panel) then performs explicit and deterministic geometric reasoning over the map, yielding a transparent and interpretable answer.} 
        \label{fig:pipeline}
\end{figure*}

%In response to 
To address 
these challenges, we propose Map2Thought, an explicit 3D spatial reasoning framework grounded in Metric Cognitive Maps that enables reliable and interpretable 3D understanding, as shown in Fig.~\ref{fig:pipeline}. Map2Thought is built upon two core components: 1) \emph{Metric-CogMap}, a unified spatial representation that combines a discrete grid for symbolic relational reasoning with a continuous metric-scale grid for precise geometric perception; and 2) \emph{Cog-CoT}, an explicit chain-of-thought reasoning module that performs interpretable geometric computations over the \emph{Metric-CogMap}.
Unlike prior cognitive maps used in~\cite{vsibench, mindcube}, our \emph{Metric-CogMap} encodes richer spatial detail, including object occupancy, metric-scale positions, and real-world scale bounding boxes—enabling more fine-grained spatial inference. Our proposed \emph{Metric-CogMap} is constructed through a robust video-to-map pipeline that extends state-of-the-art 2D object detection and segmentation models—originally designed for frame-based input and output—by integrating 3D vision foundation models and a novel covisibility map–based geometric validation step to ensure accurate and consistent 3D spatial grounding.
\emph{Cog-CoT} complements this representation by performing transparent, modular geometric reasoning (e.g., distance, direction) through deterministic computations, enabling verifiable inference traces and making our framework easily extensible to other 3D-VLMs without retraining.
With this design, Map2Thought attains $59.9\%$ accuracy on VSI-Bench using only $50\%$ of the training data—nearly matching the $60.9\%$ full-data baseline—while still exceeding the baseline by a consistent $4.0\%$ margin under equal $50\%$ supervision.
%\dave{please provide more quantified results there, i.e., achieving xxx\% performance boost with only yyy\% of the training data.}
%\dave{it would be more thrilling for the reader to see results like `reaching SOTA on zzz benchmark', e.g. VSI-Benchmark.}

\noindent The main contributions of Map2Thought are threefold:
\begin{itemize}
%
% \item \textbf{Explicit Metric-CogMap representation:} We propose a spatial abstraction that unifies symbolic reasoning and metric geometry, enabling LLMs to interpret 3D scenes with structured, interpretable reasoning.
\item \textbf{Explicit Metric-CogMap representation:} We introduce \emph{Metric-CogMap}, a unified spatial representation that integrates discrete symbolic grids with continuous metric-scale geometry, enabling LLMs to interpret 3D scenes with structured, interpretable reasoning.
\item \textbf{Cog-CoT reasoning paradigm:} We introduce \emph{Cog-CoT}, an interpretable CoT that operates over the accumulated spatial context of the Metric-CogMap to perform explicit and verifiable 3D spatial reasoning.
\item \textbf{Data-efficient 3D-VLMs:} Our framework exhibits strong data efficiency, achieving better accuracy than comparable baselines under limited training data, while improving generalization by mitigating data-dependent overfitting.
\end{itemize}

\section{Related Work}
\label{sec:related}

% Draft by EP
Large Language Models (LLMs) have progressed rapidly in recent years, with active research expanding their capabilities across diverse domains such as code generation, commonsense reasoning, and mathematics~\citep{touvron2023llama2}. A prominent direction in this evolution involves extending LLMs to handle multiple modalities—enabling them to develop visual~\citep{qwen2_5_vl} and spatial understanding~\citep{gpt4scene}. Early Visual-Language Models (VLMs), such as CLIP~\cite{radford2021clip} and ALIGN~\cite{jia2021align}, learn joint image-text embeddings, while later works~\cite{alayrac2022flamingo, li2023blip2} decouple vision and language modules to better support cross-modal reasoning. While these models perform well on single-image tasks, they remain limited in reasoning about 3D spatial context intuitive to humans—such as distance, relative positioning, and object counting.

\noindent \textbf{Point cloud encoding in 3D VLMs.}  
Recent 3D-VLMs integrate point clouds with RGB and text inputs to provide geometric grounding, but their fusion in latent space often limits spatial precision and interpretability~\citep{li2025does}. Models such as ~\citep{leo, deng20253d} use spatial transformers~\citep{chen2022vil3drel} to encode object-centric 3D tokens and capture inter-object geometry, while ~\citep{hong20233d, chen2024ll3da} align 2D and 3D features by projecting them back to reconstructed points for enhanced multimodal understanding. However, both rely on latent feature-level fusion across modalities—point clouds, images, and text—leading to opaque reasoning pipelines that hinder interpretability and analysis. More explicit pipelines, introduced by~\citep{Inst3D-LMM, chat3d, li20243dmit}, combine multi-view 2D semantics features with 3D geometric information for improved indoor scene understanding. However, they assume access to high-quality, densely reconstructed point clouds—an assumption that breaks down in real-world settings where input videos span several minutes, as is common in datasets like ScanNet\cite{dai2017scannet}, ScanNet++\cite{scannetpp} or ARKitScenes\cite{arkitscenes}. In such cases, sparse sampling or partial reconstructions may miss key objects, making reliance on complete point cloud–based instance detection a major bottleneck. %Moreover, fusion at the feature level remains largely opaque, hindering interpretability and limiting our ability to diagnose or improve the reasoning process—an ongoing challenge for early-stage 3D-VLM research.

\noindent \textbf{Positional encodings for 3D VLMs.} 
To improve spatial grounding, recent 3D VLMs introduce positional encodings derived from camera poses, depth maps, or back-projected coordinates. These encodings localize visual features within a 3D coordinate system, aiding alignment across views and with language~\citep{ma2024llms}. For instance, \textsc{LLaVA‑3D}~\citep{llava3d} and \textsc{Video‑3D‑LLM}~\citep{video3dllm} embed spatial tokens linking 2D appearances to 3D context. While this enhances object-level spatial sensitivity, they~\cite{llava3d, wang2025ross3d, video3dllm} often fail to capture broader geometric structure. Consequently, the resulting representations remain opaque and overfitting, limiting interpretability and generalization in spatial reasoning tasks.

\noindent \textbf{3D Vision foundational model encodings for 3D VLMs.} 
More recently, emerging approaches have incorporated pretrained 3D vision foundation models—such as CUT3R~\citep{cut3r} and VGGT~\citep{wang2025vggt}—into 3D-VLM pipelines to enhance spatial reasoning through learned geometric priors. For example, some recent works~\citep{fan2025vlm, vgllm, wu2025spatialmllmboostingmllmcapabilities, huang2025mllms, xu2025uniugg}, process monocular RGB video sequences using geometry encoders that produce implicit 3D tokens, which are fused with visual tokens from existing VLMs. While this fusion improves spatial feature encoding, it still relies on latent representations derived from 2D inputs—lacking explicit metric scale, object-level semantics, and coherent cross-modal alignment. As a result, the encoded spatial structures remain difficult to interpret and limit the model’s ability to perform tasks requiring precise geometric reasoning—such as appearance order, object size, or absolute distance in VSI-Bench~\citep{vsibench}.

\noindent \textbf{Cognitive maps.}  
Motivated by principles of human spatial cognition, recent works have explored cognitive map-style representations that discretize 3D environments into symbolic spatial grids encoding relative object positions and inter-object relationships. Approaches like VSI‑Bench~\citep{vsibench}, SpatialMind~\citep{zhang2025spatial}, and MindCube~\citep{mindcube} enable spatial reasoning through coarse symbolic cognitive maps based on integer grid coordinates. More recently, this direction has progressed with the integration of scene graphs \citep{cao2024cognav} to model richer relational semantics. While these symbolic representations enhance interpretability and high-level reasoning, they still lack geometric precision and omit metric details essential for fine-grained spatial understanding. 
%
% Furthermore, aligning symbolic spatial priors with noisy, ambiguous RGB inputs from subsampled video frames remains a challenge, ultimately limiting the accuracy, transparency, and scalability of spatial reasoning in 3D VLMs.

Despite recent advances, 3D-VLMs still lack explicit spatial representations—especially under noisy, subsampled inputs—limiting their ability to support interpretable and transparent reasoning. Prior work tends to prioritize benchmark performance over explainability, offering limited insight into spatial understanding. In this paper, we introduce a framework that constructs metrically grounded cognitive maps and enables chain-of-thought spatial reasoning, enhancing robustness, interpretability, and scalability.% in 3D VLMs.

\section{The Proposed Framework: Map2Thought}
\label{sec:method}

\noindent\textbf{Overview.} We introduce Map2Thought, a unified framework for 3D spatial understanding from monocular RGB videos and language queries, designed to enable explicit and interpretable reasoning grounded in structured geometric representations. As shown in Fig.~\ref{fig:pipeline}, Map2Thought integrates pretrained vision-language and 3D vision foundation models to jointly process visual, geometric, and textual inputs. The framework is composed of three key components: 1) defining the \emph{Metric-CogMap}, a dual-format representation that encodes both discrete spatial layouts and continuous metric-scale geometry to support relational reasoning and precise spatial grounding; 2) implementing a video-to-map pipeline that constructs the Metric-CogMap from raw video frames, leveraging object detection, tracking, and feed-forward 3D reconstruction; and 3) adopting \emph{Cog-CoT} (Cognitive Chain-of-Thought), a structured reasoning mechanism that performs explicit and verifiable spatial inference over the constructed \emph{Metric-CogMap}.

These components operate in two complementary streams: one for extracting and fusing multimodal tokens for contextual understanding, and another for constructing a structured scene representation (Metric-CogMap) used in explicit reasoning (Cog-CoT). These components work in tandem to produce interpretable and geometry-aware outputs. We detail each part of the architecture below.

%The remainder of this section is organized as follows. Sec.~\ref{sec:architecture} provides an overview of the paper pipeline. Sec.~\ref{sec:cogmap-design} details the design principles and structure of the \emph{Metric-CogMap} representation. Sec.~\ref{sec:cogmap-construction} describes the extraction process from video input. Finally, Sec.~\ref{sec:cog-cot} presents our \emph{Cog-CoT} reasoning framework.

% ==================================
% Overall architecture illustration
% ==================================
\subsection{Architecture}
\label{sec:architecture}
%\noindent\textbf{Overview.} The Map2Thought framework, shown in Fig.~\ref{fig:pipeline}, performs 3D spatial reasoning from monocular RGB video and a language query. It first extracts 2D image tokens and 3D geometry-aware tokens using a pretrained VLM image encoder~\citep{zhang2024llavanextvideo} and 3D vision foundation models~\citep{cut3r}. The text query is tokenized and processed jointly with these visual tokens by the VLM~\citep{zhang2024llavanextvideo}. In parallel, a \emph{Metric-CogMap} is constructed using object detection~\citep{zhou2022detecting, liu2024grounding}, video segmentation~\citep{ravi2025sam2}, and feed-forward 3D scene reconstruction modules~\citep{wang2025pi3} applied to the input frames. On top of this map, \emph{Cog-CoT} performs query-driven, deterministic reasoning steps, enabling explicit and interpretable geometric inference grounded in scene structure.
%\noindent\textbf{Overview.}
%Map2Thought is a unified framework designed for explicit 3D spatial reasoning from monocular videos and language queries. As illustrated in Fig.~\ref{fig:pipeline}, it integrates pretrained vision-language and 3D foundation models to jointly process visual, geometric, and textual inputs. The framework builds upon two complementary streams: one for extracting and fusing multimodal tokens for contextual understanding, and another for constructing a structured scene representation used in explicit reasoning. These components work in tandem to produce interpretable and geometry-aware outputs. We describe each in detail below.

\begin{figure}%[htpb] % The placement creates weird problems (image on second line). Unless there are good reasons, leave latex to choose the placement.
	\centering
	\includegraphics[width=1.0 \linewidth]{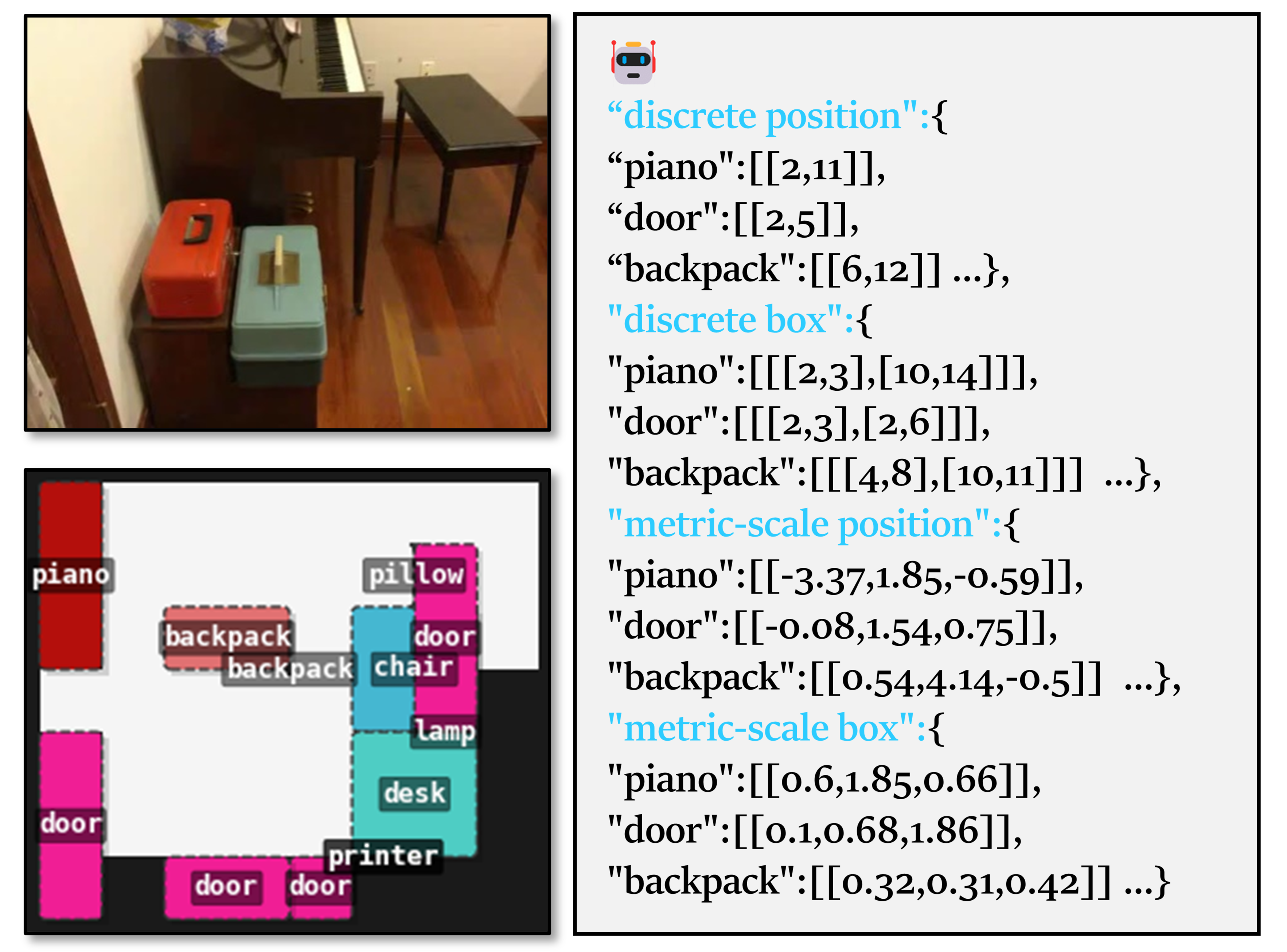}
        \caption{Visualization of a \emph{Metric-CogMap} example. Given the reference frame shown at the top left, the discrete grid representation (top two) assigns each object an integer-valued position, while the box occupancy map encodes the grid ranges each object occupies. The metric-scale representation (bottom two) records object centroids and axis-aligned bounding boxes (AABBs) in a globally aligned real-world coordinate system.
        }
        \label{fig:cogmap}
\end{figure}

%\smallskip\noindent\textbf{Visual and geometry token} 
%We adopt a 2D VLM backbone (e.g., LLaVA-Next-Video). A frozen visual encoder (e.g., CLIP ViT) processes each resized video frame (e.g., $432{\times}432$) to produce 2D visual tokens $Z_{2D}$ that capture appearance and local spatial cues. In parallel, a frozen geometric stream (e.g., CUT3R~\cite{cut3r}) provides 3D reconstructive tokens that encode global scene geometry across the sequence. We concatenate them into a unified 3D token representation $Z_{3D}$ with a fixed token length per frame, ensuring compatibility with the visual stream for subsequent fusion.

%\smallskip\noindent\textbf{Spatial–visual token fusion.} Similar to VLM-3R\cite{vlm3r}, we fuse the 2D visual branch with the 3D reconstructive branch via a lightweight cross-attention module, where the visual tokens $Z_{2D}$ serve as queries and the 3D tokens $Z_{3D}$ serve as keys and values. This operation produces geometry‐enriched visual tokens $Z_f$ that integrate appearance cues with global spatial structure. A two‐layer projector~\cite{zhang2024llavanextvideo} is then applied to map $Z_f$ into the latent space of the VLM. 
%%Finally, the fused token sequence $[Z_f; \text{text}]$
%\zs{Finally, the fused token sequence $[Z_f; Z_t]$ of the visual token and text token}
%is passed to the VLM’s transformer for subsequent spatial task reasoning.

\noindent\textbf{Token extraction and fusion.}
Map2Thought processes a monocular RGB video and a text query to perform 3D spatial reasoning. Each video frame is resized and passed through a frozen 2D visual encoder (e.g., CLIP-ViT~\citep{radford2021clip}) to extract 2D appearance tokens $Z_{2D}$, capturing local semantic and visual cues. Simultaneously, a pretrained 3D vision foundation model (e.g., CUT3R~\citep{cut3r}) produces geometry-aware tokens $Z_{3D}$ that encode the global scene structure across frames. These 2D and 3D tokens are fused using a lightweight cross-attention mechanism~\citep{vlm3r}, where $Z_{2D}$ queries $Z_{3D}$ to generate geometry-enriched visual tokens $Z_f$. The fused tokens are then projected into the VLM’s latent space and concatenated with the tokenized language query for downstream spatial reasoning.

%\smallskip\noindent\textbf{Metric-CogMap and Cog-CoT.}
%Beyond spatial–visual fusion, we introduce two components enabling explicit and verifiable spatial reasoning. The \emph{Metric-CogMap} is constructed from the video stream and maintains a dual structured representation that combines discrete spatial layouts with continuous metric-scale geometry, capturing object locations, extents, and inter-object relationships in an interpretable form. In parallel, \emph{Cog-CoT} (Cognitive Chain-of-Thought) performs explicit and interpretable reasoning through deterministic geometric computations that are \emph{grounded in} the \emph{Metric-CogMap}. In doing so, Cog-CoT yields a verifiable spatial reasoning process. This joint design enables the system to produce spatially accurate outputs while retaining transparent and interpretable reasoning trajectories.

\noindent\textbf{Metric-CogMap and Cog-CoT.}
In parallel to token fusion, Map2Thought constructs a \emph{Metric-CogMap} from the input video using off-the-shelf modules for object detection~\citep{zhou2022detecting}, video segmentation~\citep{ravi2025sam2}, and feed-forward 3D reconstruction~\citep{wang2025pi3}. This map encodes both discrete spatial grids and continuous metric-scale object geometry, including positions, bounding boxes, and relational layouts. Built atop this representation, \emph{Cog-CoT} (Cognitive Chain-of-Thought) performs explicit reasoning through deterministic operations (e.g., vector relations, occlusion-aware queries) grounded in the map. Together, these modules enable transparent and verifiable 3D reasoning that enhances both accuracy and interpretability under varied supervision regimes.

% ==================================
% How we design the Metric-CogMap
% ==================================
\subsection{Design of Metric-CogMap}
\label{sec:cogmap-design}

\begin{figure*}[htpb]
	\centering
	\includegraphics[width=1.0 \linewidth]{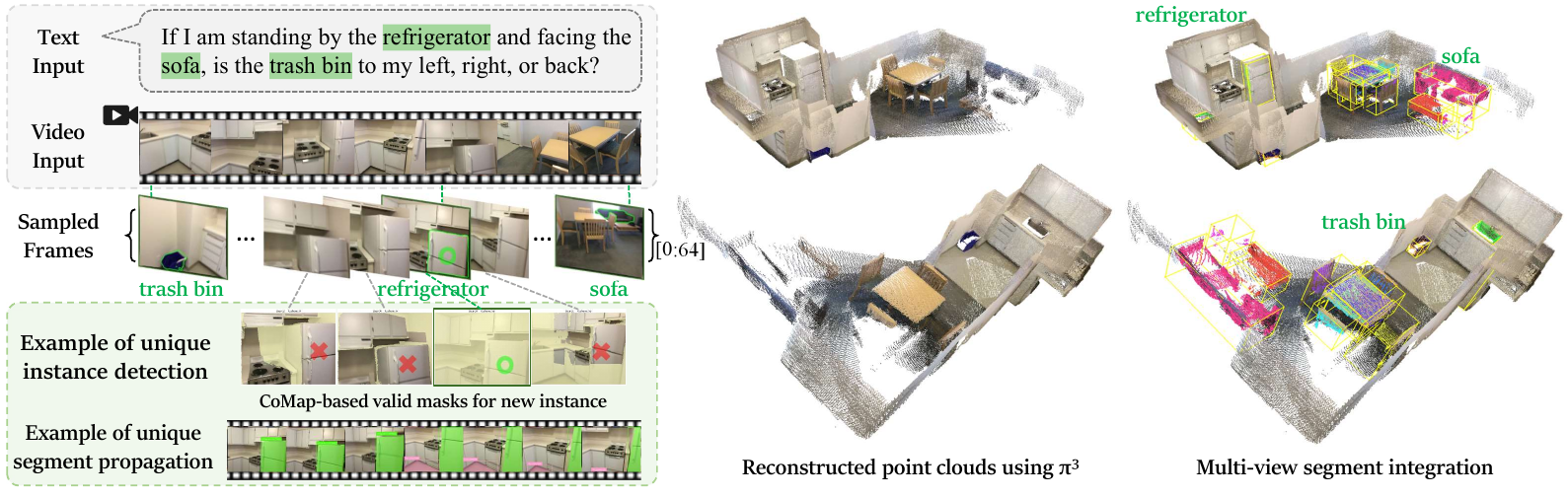}
        \caption{\textbf{Overview of the Metric-CogMap construction pipeline.} Given video and text inputs, the pipeline selects 64 uniformly distributed frames while ensuring key frames containing the highest-confidence detections of text-referenced objects (green outlines and annotations). For categories requiring multi-instance detection (e.g., the refrigerator), covisibility maps anchor the highest-confidence frame and suppress redundant detections by checking whether candidate regions appeared in earlier selected views. Only geometrically unique detections within valid mask regions are retained. These detections are propagated across frames, and multi-view segments are merged with consistent IDs during final integration. The example on the right shows the resulting detections for several question-relevant categories (e.g., sofa, trash bin, chair, table) from \texttt{scene0651\_02} in the ScanNet dataset~\citep{dai2017scannet}, an evaluation scene in VSI-Bench~\citep{vsibench}.}  
        \label{fig:metric_cogmap_pipeline}
\end{figure*}

Our \emph{Metric-CogMap} provides a unified spatial representation that integrates two complementary components—(i) a discrete grid-based abstraction and (ii) a continuous metric-scale encoding.
Together, these representations support structured, interpretable, and verifiable spatial reasoning by combining symbolic relational structure with precise geometric information.

\smallskip\noindent\textbf{Discrete grid representation.}
First, the scene is represented on a \emph{discrete position representation}, where each object is assigned an integer coordinate (e.g., within a fixed range such as 0–20).
Beyond mapping an object to a single cell, we also encode its occupied region using four integer values that specify an axis-aligned bounding box, forming a \emph{discrete box representation} that preserves its spatial extent.
This discrete encoding offers an approximate but structured description of position and size, enabling symbolic reasoning and relational parsing in a simplified grid domain while reducing the complexity of spatial inference.

\smallskip\noindent\textbf{Continuous metric-scale representation.}
In parallel, we preserve real-world spatial fidelity by annotating each object with its \emph{metric-scale position} in the real-world coordinate system.
We further associate each object with a \emph{metric-scale axis-aligned bounding box (AABB)}, which provides its physical size and exact spatial boundaries in continuous 3D space.
This continuous formulation retains true geometric structure and distance relationships, enabling precise spatial reasoning that discrete layouts cannot capture.

\smallskip\noindent\textbf{Unified representation.}
Fig.~\ref{fig:cogmap} illustrates the integrated \emph{Metric-CogMap}.
The discrete grid representation (top two panels) assigns each object an integer-valued position and encodes its occupied grid range through a discrete bounding box, yielding a symbolic and interpretable layout of the scene.
Complementarily, the metric-scale representation (bottom two panels) records each object's real-world centroid and its physical axis-aligned bounding box (AABB), capturing precise geometric size and spatial extent in the global coordinate frame.
By coherently linking these \emph{discrete} and \emph{continuous}  components, the unified map provides a structured relational view aligned with precise geometric measurements, while maintaining an explicit and interpretable inference process.

% ==================================
% How we construct the Metric-CogMap
% ==================================
\subsection{Construction of Metric Cognitive Map}
\label{sec:cogmap-construction}

To support spatially grounded reasoning in our proposed Map2Thought framework, we construct a Metric-CogMap through a multi-stage pipeline that integrates visual semantics with metrically aligned 3D geometry, as shown in Fig.~\ref{fig:metric_cogmap_pipeline}. The process involves: 1)~Crucial frame selection via key-object anchoring, 2) covisibility map construction, 3) multi-instance-aware object detection (conditional), 4) object-centric video segmentation, 5) geometric transformation and metric alignment, and 6) multi-view segment integration.

\noindent \textbf{Crucial frame selection via key-object anchoring}: To ensure that all objects referenced in the input question are visually represented in the selected frames, we replace naïve uniform subsampling with a targeted frame selection strategy. We first apply the Detic detector~\citep{zhou2022detecting} to every 10th frame to identify high-confidence detections of queried objects, retaining the most representative frame for each as a crucial view. To complete a 64-frame input set, we then add uniformly sampled frames from the remaining video to ensure broad spatio-temporal coverage. We then apply Grounding DINO~\citep{liu2024grounding}, an open-vocabulary object detector, to all selected frames to capture additional object instances that may have been missed by Detic—particularly for underrepresented or category-agnostic entities. This procedure guarantees that key semantic cues are explicitly embedded in the visual stream, overcoming a fundamental limitation of conventional uniform sampling that risks excluding critical objects from the reconstruction pipeline.

\noindent \textbf{Covisibility map construction}: To establish reliable inter-frame geometric correspondence, we generate dense pixel-wise Covisibility Maps (CoMaps), following the principles of CoMapGS~\citep{Jang_2025_CVPR}. These maps indicate which regions in one frame are geometrically visible from others, forming the basis for robust multi-view reasoning. Using $\pi^3$~\citep{wang2025pi3}, we extract dense 3D point clouds and their associated camera poses. Covisibility is then determined via reprojection consistency: a 3D point observed in a source frame is projected into target views, and covisibility is confirmed if the projection aligns with neighboring 3D structure in the target frame. These pairwise pixel-level covisibility cues are essential for subsequent stages, supporting accurate multi-view fusion, disambiguation of duplicate object detections, and spatial validation of instance consistency.

\noindent \textbf{Multi-instance-aware object detection (conditional)}: For queries requiring multi-instance understanding—such as object counting or appearance ordering—we introduce an additional detection refinement step guided by geometric constraints. Leveraging the previously constructed CoMaps, we enforce spatial distinctiveness by filtering out redundant detections across views. For each object category, we start with the highest-confidence detection and iteratively suppress other candidates whose masks fall within the covisible regions of already-selected instances. This strategy ensures non-redundant, viewpoint-distinct detections. This deduplication process is applied across both Detic and Grounding DINO~\citep{liu2024grounding} outputs, ensuring that only spatially distinct instances are preserved in the map.

\noindent \textbf{Object-centric video segmentation}: To enhance both the geometric integrity and semantic completeness of object representations, we apply SAM2~\citep{ravi2025sam2} to propagate detected instances across frames.This enables the recovery of more complete object regions beyond the initial detection frame, providing a more holistic understanding of each object’s shape and spatial extent.

\noindent \textbf{Geometric transformation and metric alignment}: Leveraging the camera poses and 3D point clouds provided by $\pi^3$~\citep{wang2025pi3}, we perform global alignment by anchoring the reconstruction to a dominant horizontal plane (e.g., the floor), or a vertical planewhen it provides a more reliable reference. To achieve metric consistency, we rescale the reconstruction based on depth from MoGe-2~\citep{wang2025moge2accuratemonoculargeometry}, producing a metrically accurate, globally aligned 3D scene. 

% This alignment is essential for supporting downstream spatial reasoning tasks that require interpretable and geometrically reliable structure.
% why we need this commented sentence? 

\noindent \textbf{Multi-view segment integration}: Finally, segmented object instances and their associated 3D points are consolidated across views to form a unified spatial representation. Overlapping regions—defined as 3D segments within a $20\%$ spatial proximity threshold—are merged by retaining the highest-confidence detection, effectively reducing redundancy and addressing potential false duplicates that may still arise from conventional detection pipelines. Merged objects are then assigned consistent semantic labels and unique instance IDs, producing a compact, semantically coherent, and metrically grounded \emph{Metric-CogMap}.

This integrated map encodes both object-level semantics and spatial topology, providing a robust foundation for structured spatial reasoning via \emph{Cog-CoT} in 3D VLMs.

\subsection{Cog-CoT: Cognitive Chain-of-Thought}
\label{sec:cog-cot}

Our \emph{Cog-CoT} (Cognitive Chain-of-Thought) enables explicit and verifiable spatial reasoning—unlike prior approaches that rely on end-to-end predictions over visual inputs with implicit, non-interpretable processes. It operates directly on the structured geometry encoded in the \emph{Metric-CogMap}, executing deterministic operations such as vector analysis, bounding-box distance, and appearance order. This structured reasoning chain allows the VLM to apply its language understanding to interpret geometric evidence and derive the final answer. In the following, we demonstrate how Cog-CoT applies these operations across representative categories of spatial queries, enabling transparent and interpretable 3D reasoning.
% \emph{Cog-CoT} decomposes spatial queries into sequences of deterministic mathematical computations that are fully grounded in the \emph{Metric-CogMap}, ensuring interpretability and verifiability.

\begin{figure}[htpb]
	\centering
	\includegraphics[width=1.0 \linewidth]{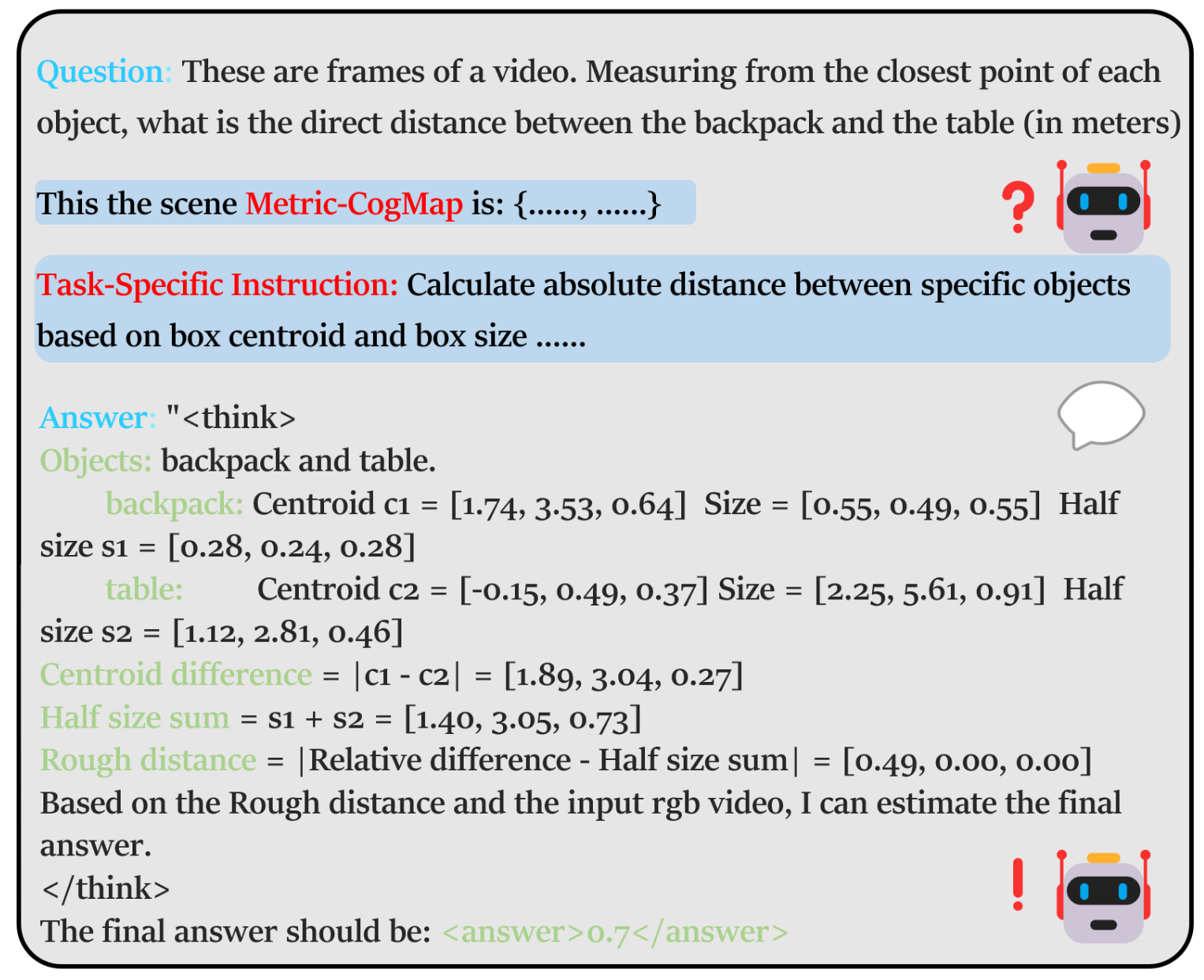}
        \caption{\textbf{Cog-CoT workflow.} Given a query, the \emph{Metric-CogMap}, and the corresponding \emph{Task-Specific instruction}, the system retrieves the scene-level \emph{Metric-CogMap}, and \emph{Cog-CoT} executes the geometric computations step-by-step, yielding verifiable intermediate numerical evidence used to generate the final answer.}
        \label{fig:cogcot}
\end{figure}

\begin{table*}[ht!] % Using table* environment
    \captionsetup{type=table}
    \centering
    % \begin{minipage}{0.83\textwidth}
    \centering
    \fontsize{3.6pt}{3.4pt}\selectfont % Keeping original font size
    \setlength\tabcolsep{3pt} % Keeping tabcolsep
    \renewcommand{\arraystretch}{1.2}
    % \scalebox{1.57}{
    \resizebox{0.9\textwidth}{!}{
    \begin{tabular}{r|cc|cccccccc}
    % --- Task Names Row ---
    & & & % Methods, Rank, Avg columns
    \rotatebox{70}{Obj. Count} &
    \rotatebox{70}{Abs. Dist.} &
    \rotatebox{70}{Obj. Size} &
    \rotatebox{70}{Room Size} &
    \rotatebox{70}{Rel. Dist.} &
    \rotatebox{70}{Rel. Dir.} &
    \rotatebox{70}{Route Plan} &
    \rotatebox{70}{Appr. Order} \\
    % --- Task Types Row ---
    Methods & Rank & Avg. &
    \multicolumn{4}{c}{\cellcolor{orange!30}Numerical Answer} &
    \multicolumn{4}{c}{\cellcolor{yellow!30}Multiple-Choice Answer} \\
    \hline
    % --- Baseline Section ---
    \rowcolor{navyblue!5}
    \multicolumn{11}{l}{\textcolor{black}{\textit{Baseline}}} \\
    Chance Level (Random) & - & - & - & - & - & - & 25.0 & 36.1 & 28.3 & 25.0 \\
    Chance Level (Frequency) & - & 34.0 & 62.1 & 32.0 & 29.9 & 33.1 & 25.1 & 47.9 & 28.4 & 25.2 \\
    \hline
    % \rowcolor{navyblue!5}
    % \multicolumn{11}{l}{\textcolor{black}{\textit{VSI-Bench Perf. (\dag = Tiny Set)}}} \\
    % \dag Human Level & - &79.2 &94.3 &47.0 &60.4 &45.9 &94.7 &95.8 &95.8 &100.0 \\
    % \dag Gemini-1.5 Flash & - & 45.7 & 50.8 & 33.6 & 56.5 & 45.2 & 48.0 & 39.8 & 32.7 & 59.2 \\
    % \dag Gemini-1.5 Pro & - & 48.8 & 49.6 & 28.8 & 58.6 & 49.4 & 46.0 & 48.1 & 42.0 & 68.0 \\
    % \dag Gemini-2.0 Flash & - & 45.4 & 52.4 & 30.6 & 66.7 & 31.8 & 56.0 & 46.3 & 24.5 & 55.1 \\
    % \hline
    % \rowcolor{navyblue!5}
    \rowcolor{cyan!10}
    \multicolumn{11}{l}{\textcolor{black}{\textit{Proprietary Models (API)}}} \\
    GPT-4o & \cellcolor{oai-green-200}{3} & 34.0 & 46.2 & 5.3 & 43.8 & 38.2 & 37.0 & 41.3 & 31.5 & 28.5 \\
    Gemini-1.5 Flash & \cellcolor{oai-green-400}{2} & 42.1 & 49.8 & 30.8 & 53.5 & 54.4 & 37.7 & 41.0 & 31.5 & 37.8 \\
    Gemini-1.5 Pro & \cellcolor{oai-green-600}{1} & 45.4 & 56.2 & 30.9 & 64.1 & 43.6 & 51.3 & 46.3 & 36.0 & 34.6 \\
    \hline
    \rowcolor{cyan!10}
    % \rowcolor{navyblue!5}
    \multicolumn{11}{l}{\textcolor{black}{\textit{Open-sourced VLMs}}} \\
    % Sub-group: Small Models (0.5B-2B)
    LLaVA-OneVision-0.5B & 16 & 28.0 & 46.1 & 28.4 & 15.4 & 28.3 & 28.9 & 36.9 & 34.5 & 5.8 \\
    InternVL2-2B & 17 & 27.4 & 21.8 & 24.9 & 22.0 & 35.0 & 33.8 & {44.2} & 30.5 & 7.1 \\ % Rel.Dir: 2nd OS
    \hline % Separator
    % Sub-group: Medium Models (7B-8B, excluding VLM-3R)
    LLaVA-NeXT-Video-7B & 10 & 35.6 & 48.5 & 14.0 & 47.8 & 24.2 & {43.5} & 42.4 & 34.0 & 30.6 \\ % Rel.Dist: 2nd OS
    InternVL2-8B & 11 & 34.6 & 23.1 & {28.7} & 48.2 & {39.8} & 36.7 & 30.7 & 29.9 & 39.6 \\ % Abs.Dist: 2nd OS; Room Size: 2nd OS
    LLaVA-OneVision-7B & 12 & 32.4 & 47.7 & 20.2 & 47.4 & 12.3 & 42.5 & 35.2 & 29.4 & 24.4 \\
    LongVA-7B & 14 & 29.2 & 38.0 & 16.6 & 38.9 & 22.2 & 33.1 & 43.3 & 25.4 & 15.7 \\
    VILA-1.5-8B & 15 & 28.9 & 17.4 & 21.8 & 50.3 & 18.8 & 32.1 & 34.8 & 31.0 & 24.8 \\
    LongVILA-8B & 18 & 21.6 & 29.1 & 9.1 & 16.7 & 0.0 & 29.6 & 30.7 & 32.5 & 25.5 \\
    \hline % Separator
    % Sub-group: Large Models (40B)
    InternVL2-40B & 9 & 36.0 & 34.9 & 26.9 & 46.5 & 31.8 & 42.1 & 32.2 & 34.0 & 39.6 \\
    VILA-1.5-40B & 13 & 31.2 & 22.4 & 24.8 & 48.7 & 22.7 & 40.5 & 25.7 & 31.5 & 32.9 \\
    % \hline % Separator
    % Sub-group: Extra Large Models (72B)
    LLaVA-NeXT-Video-72B & {7} & 40.9 & {48.9} & 22.8 & 57.4 & 35.3 & 42.4 & 36.7 & {35.0} & {48.6} \\ % Obj.Count: 2nd OS; Route Plan: 2nd OS; Appr.Order: 1st OS (also overall best)
    LLaVA-OneVision-72B & {8} & 40.2 & 43.5 & 23.9 & {57.6} & 37.5 & 42.5 & 39.9 & 32.5 & {44.6} \\ % Obj.Size: 2nd OS; Appr.Order: 2nd OS
    
    % baseline Model
    \hline 
    Spatial-MLLM-4B & 6 & 48.4 & 65.3 & 34.8 & 63.1 & 45.1 & 41.3 & 46.2 & 33.5 & 46.3 \\
    VG-LLM-8B & 4 & 58.2 & \cellcolor{oai-gray-600}{71.7} & \cellcolor{oai-gray-300}{53.8} & \cellcolor{gray!10}{68.8} & 62.1 & \cellcolor{oai-gray-300}{63.8} & \cellcolor{oai-gray-600}{83.0} & \cellcolor{oai-gray-300}{44.3} & 18.4 \\
    
    % VLM-3R and ours 100%
    \hline 
    {VLM-3R (100\% dataset)} & \cellcolor{oai-green-400}{2} & 60.9 & \cellcolor{gray!10}{70.2} & {49.4} & \cellcolor{oai-gray-300}{69.2} & \cellcolor{gray!10}{67.1} & \cellcolor{oai-gray-600}{65.4} & \cellcolor{oai-gray-300}{80.5} & \cellcolor{oai-gray-600}{45.4} & 40.1 \\ 
    % \textbf{VLM-3R 100\% dataset} & \cellcolor{oai-green-600}{1} & 59.94 & \cellcolor{oai-gray-600}{70.8} & \cellcolor{oai-gray-600}{49.1} & \cellcolor{oai-gray-600}{70.8} & \cellcolor{oai-gray-600}{66.2} & \cellcolor{oai-gray-600}{63.3} & \cellcolor{oai-gray-600}{74.3} & \cellcolor{oai-gray-600}{43.8} & 40.9 \\
    % {Ours (100\% dataset)} &  \cellcolor{oai-green-600}{1} & 60.97 & \cellcolor{oai-gray-600}{70.88} & \cellcolor{oai-gray-600}{55.02} & \cellcolor{oai-gray-600}{70.16} & \cellcolor{oai-gray-600}{69.41} & {56.90} & {69.80} & {38.14} & \cellcolor{oai-gray-600}{57.44} \\
    \textbf{Ours (100\% dataset)} &  \cellcolor{oai-green-600}{1} & 61.0 & \cellcolor{oai-gray-300}{70.8} & \cellcolor{oai-gray-600}{55.0} & \cellcolor{oai-gray-600}{70.1} & \cellcolor{oai-gray-600}{69.4} & {56.9} & {69.8} & {38.1} & \cellcolor{oai-gray-600}{57.4} \\

    % VLM-3R and ours 25%
    \hline
    % \textbf{VLM-3R (50\% dataset)} & rank & 55.54 & 70.14 & 43.74 & 69.77 & 61.49 & 58.59 & 52.58 & 43.81 & 44.17 \\
    % \textbf{Ours (50\% dataset)} & rank & 59.53 & 70.23 & 54.27 & 66.60 & 68.82 & 55.49 & 69.01 & 37.11 & 54.69 \\
    % \hline
    {VLM-3R (25\% dataset)} & 5 & 54.0 & 69.2 & 41.7 & 67.6 & 62.4 & \cellcolor{gray!10}{58.4} & 47.3 & \cellcolor{gray!10}{43.3} & \cellcolor{gray!10}{41.7} \\
    \textbf{Ours (25\% dataset)} & \cellcolor{oai-green-200}{3} & 58.8 & 68.2 & \cellcolor{gray!10}{53.0} & 64.4 & \cellcolor{oai-gray-300}{69.1} & 52.1 & \cellcolor{gray!10}{70.9} & 38.1 & \cellcolor{oai-gray-300}{54.5} \\
    % \textbf{Ours (25\% dataset)} & \cellcolor{oai-green-200}{3} & 58.81 & 68.28 & \cellcolor{oai-gray-300}{53.05} & 64.39 & \cellcolor{oai-gray-300}{69.10} & 52.11 & \cellcolor{oai-gray-300}{70.91} & 38.14 & \cellcolor{oai-gray-300}{54.53} \\
    % \hline
    % \textbf{VLM-3R (10\% dataset)} & rank & 49.12 & 67.89 & 37.37 & 64.91 & 53.54 & 53.38 & 47.00 & 37.11 & 31.72  \\
    % \textbf{Ours (10\% dataset)} & rank & 54.51 & 61.72 & 51.52 & 59.62 & 69.72 & 41.97 & 70.09 & 26.29 & 55.18 \\

    \end{tabular}
    } % end scalebox
    % \end{minipage}
    \caption{\textbf{Evaluations on VSI-Bench.} Our model ranks first among open-sourced VLMs, showcasing the effectiveness of \emph{Metric-CogMap} and \emph{Cog-CoT}.  For each task within the open-sourced 3D-VLM group, the \colorbox{oai-gray-600}{dark gray} cell indicates the best-performing model, while \colorbox{oai-gray-300}{light gray} and \colorbox{gray!10}{faint gray} highlight the second- and third-best performances, respectively. }
    \label{tab:vsibench}
    \vspace{-3mm}
\end{table*}

%Fig.~\ref{fig:cogcot} illustrates a representative reasoning trace for the task of \emph{Absolute Distance Measurement}. It shows the question (requesting the direct distance between two selected objects), the task-specific instruction (providing guidance on how to answer the question), and the scene’s \emph{Metric-CogMap} (as depicted in Fig.~\ref{fig:cogmap}). The VLM model retrieves the relevant information from the \emph{Metric-CogMap}, with only the metric-scale representation needed in this case to answer the question. The VLM model then performs explicit geometric operations, such as box size estimation and distance calculation, to derive verifiable intermediate results before generating the final answer.

Fig.~\ref{fig:cogcot} illustrates a representative \emph{Cog-CoT} trace for the \emph{Absolute Distance Measurement} task: \\
\noindent\textbf{Query and instruction:} The language query requests the metric distance between two specified objects, paired with a task-specific instruction that guides the expected reasoning procedure. \\
\noindent\textbf{Scene representation:} The relevant \emph{Metric-CogMap} (see Fig.~\ref{fig:cogmap}) is provided, with the metric-scale representation used to retrieve object centroids and bounding boxes. \\
\noindent\textbf{Geometric reasoning:} The model applies explicit geometric operations (e.g., centroid extraction, distance calculation) to generate verifiable intermediate results. \\
\noindent\textbf{Answer generation:} These intermediate results guide the final response in a transparent, verifiable manner.

In addition to the distance measurement example, another \emph{Cog-CoT} application—\emph{Relative Direction Reasoning}—is shown in Fig.~\ref{fig:pipeline}, involving geometric operations such as cross and dot products to determine directional relationships. Further examples and task-specific instructions are provided in the supplementary material.

\section{Experiments}
\label{sec:experiment}

% In this section, we present ...... Section \ref{}....
% \subsection{Training Details.}

% We adopt the LLaVA-Next-Video-7B as our base model and perform fine-tuning on VLM-3R data. Efficient adaptation is achieved with LoRA~\cite{hu2022lora}, updating only the 2D–3D fusion attention and projection layers, strictly following the setup in VLM-3R\cite{vlm3r}. 

% \subsection{Training Dataset construction}
% Our training dataset is based on the training dataset curated by VLM-3R \cite{vlm3r} from Scannet\cite{dai2017scannet}, Scannet++\cite{scannetpp}, and ArkitScenes\cite{arkitscenes}.

% In addition to the VLM-3R data, we built ground truth metric-cogmap data for each question-answer pair.

% (1) Select the relevant object in each question.
% (2) Choose the relevant object with semantic labels and instance labels from 
% (3) The indoor scene is uniformly divided into 20*20 grid based on the room range

In this section, we evaluate our framework in terms of spatial reasoning accuracy and data efficiency. Sec.~\ref{sec:training_details} outlines the training setup, and Sec.~\ref{sec:dataset_construction} introduces the scene-level Metric-CogMap annotations. Benchmark tasks and evaluation metrics are presented in Sec.~\ref{sec:eval}, followed by an ablation study in Sec.~\ref{sec:ablation}.

\subsection{Training Details}
\label{sec:training_details}
We adopt LLaVA-Next-Video-7B as the backbon, initializing all visual and language components from publicly available checkpoints. Fine-tuning is conducted on the VLM-3R training set, strictly following the same optimization and data protocols from~\cite{vlm3r}. To ensure efficient adaptation, we employ LoRA~\cite{hu2022lora} to finetune the VLM model. We also employ the same 2D-3D fusion module and projection layer architecture as used in VLM-3R~\cite{vlm3r}.

\subsection{Training Dataset Construction}
\label{sec:dataset_construction}
Our training process builds upon the VLM-3R~\cite{vlm3r} instructional dataset, which is curated from ScanNet~\cite{dai2017scannet}, ScanNet++~\cite{scannetpp}, and ARKitScenes~\cite{arkitscenes}. For each scene and corresponding question-answer pair, we additionally construct a ground-truth \emph{Metric-CogMap}, enabling explicit spatial computation during Cog-CoT inference. The construction pipeline proceeds as follows:

\begin{enumerate}[label=(\arabic*)]
\item \textbf{Object selection.} We extract the referenced object(s) in each question via text parsing and entity grounding.
\item \textbf{Instance association.} Leveraging scene-level semantic and instance annotations, we locate the corresponding 3D object instances and obtain their axis-aligned bounding boxes (AABBs).
\item \textbf{Metric-CogMap construction.} Each indoor scene is normalized to its room extents and discretized into a $20 \times 20$ spatial grid. For each identified object, we store both a) its discrete position and occupancy on the grid and b) its continuous metric-scale position and AABB parameters in world coordinates, forming the dual-format representation.
\end{enumerate}

% This dual-form representation preserves coarse relational structure (grid layout) and precise metric geometry (AABBs), providing a direct operational substrate for explicit spatial reasoning steps executed in Cog-CoT.

\subsection{Evaluations}
\label{sec:eval}

\paragraph{Comparison baselines}
Following the evaluation protocol of VSI-Bench~\cite{vsibench}, we adopt a range of existing video LLM as baselines. These include proprietary models (e.g., Gemini~\cite{gemini_pro}, GPT\textendash4o~\cite{gpt4o}) as well as open-source counterparts such as InternVL2~\cite{chen2024internvl2}, ViLA~\cite{vila}, LongViLA~\cite{zhang2024longva}, LongVA~\cite{zhang2024longva}, LLaVA-OneVision~\cite{llava_onevision}, and LLaVA-NeXT-Video~\cite{zhang2024llavanextvideo}. The performance score of these baselines are copied from the VSI-Bench~\cite{vsibench} leaderboard.

In addition, we compare against closely related RGB-only spatial reasoning approaches, including VLM-3R~\cite{vlm3r}, VG-LLM~\cite{vgllm}, and Spatial-MLLM~\cite{wu2025spatialmllmboostingmllmcapabilities}. These methods enhance video-based LMMs by introducing tokens from 3D foundation models to improve spatial understanding. 

\paragraph{Comparison on VSI-Bench}
Our model outperforms all existing baselines, including proprietary models such as GPT-4o and Gemini-1.5-Pro, despite using only a 7B-parameter open-source backbone. We achieve state-of-the-art (SOTA) results not only in the overall average performance but also in several individual question categories, including \emph{Obj.Count}, \emph{Abs.Dist}, \emph{Obj.Size}, \emph{Room Size}, and \emph{Appr.Order}.
These gains stem from the metric-scale representation in \emph{Metric-CogMap} and geometric reasoning via \emph{Cog-CoT}. Our model excels in tasks like \emph{Abs.Dist} and \emph{Room Size}, where precise real-world understanding is essential and discrete grid maps typically fall short. By combining structured scene representations with interpretable logic, it enables accurate and reliable spatial understanding.

% In addition, our model also exhibits high data efficiency. Specifically, with 25\% of training data, we can achieve much better performance in questions that need relational reasoning, such as \emph{Rel. Dir, }.

In addition, with the proposed \emph{Metric-CogMap} and \emph{Cog-CoT}, our model enables efficient fine-tuning through a structured, interpretable spatial reasoning process. As shown in the last row of Table~\ref{tab:vsibench}, using only 25\% of the VLM-3R dataset QA pairs, it outperforms VLM-3R trained on the same subset by 4.8\% and even surpasses VG-LLM-8B trained on the full dataset. These results underscore not only the data efficiency of our method, but also how explicit metric-scale reasoning contributes to more robust generalization under limited supervision.

Although our model shows slightly lower performance on \emph{Rel.Dist} and \emph{Rel.Dir}, this is primarily due to imperfections in the constructed \emph{Metric-CogMap}, which leads to relatively lower performance in these tasks. To further validate the effectiveness of our \emph{Metric-CogMap} and \emph{Cog-CoT}, we conduct additional evaluations using the ground truth \emph{Metric-CogMap} as input, which are discussed in the subsequent ablation section.

\subsection{Ablation Study}
\label{sec:ablation}

\paragraph{Ablation on dataset proportion} 
We perform ablation studies to investigate the effect of training data proportion on model performance, comparing our approach with VLM-3R. As shown in Tab. \ref{tab:data_frac} and Fig. \ref{fig:teaser}, our model consistently outperforms VLM-3R across all dataset proportions, outperforms state-of-the-art methods by $5.3\%$, $4.8\%$, and $4.0\%$ under $10\%$, $25\%$, and $50\%$ training subsets, respectively, on the VSI-Bench. 
When trained on the full 100\% dataset, our model achieves performance comparable to VLM-3R (61.0\% vs 60.9\%).
With only 25\% of the training data, our model outperforms the VG-LLM-8B model, which is trained on 100\% of the VLM-3R training data. These improvements prove that we can achieve higher performance with fewer training data, demonstrating the efficiency and effectiveness of our approach.

We further analyze two representative tasks: \emph{Abs.Dist} and \emph{Rel.Dir}. With only 10\% of the training data, our model exceeds VLM-3R by 13.8\% and 23.0\% on these tasks, respectively. These results demonstrate that the interpretable structure of \emph{Metric-CogMap} and \emph{Cog-CoT} accelerates convergence and enhances performance, especially in geometry-sensitive tasks.

% \begin{table}[htbp]
% \centering
% \small
% \caption{Ablation study across varying training data fractions. \textbf{Bold} values indicate superior results.}
% \label{tab:data_frac}
% \setlength{\tabcolsep}{0.8 mm}{
% \begin{tabular}{lccc}
% \toprule
% \multicolumn{1}{c}{\multirow{1}{*}{Method}} & Average & Abs. Dist & Rel. Dir \\
% \midrule
% Dataset 100\% (VLM-3R) & 60.9 & 49.4 & \textbf{80.5}  \\
% Dataset 100\% (Ours) & \textbf{61.0} & \textbf{55.0} & 69.8  \\
% % Dataset 100\% (Ours) & 60.97 & 55.02 & 69.80  \\
% \midrule
% Dataset 50\% (VLM-3R) & 55.5 & 43.7 & 52.5 \\
% Dataset 50\% (Ours) & \textbf{59.5} & \textbf{54.2} & \textbf{69.0} \\
% % Dataset 50\% (Ours) & 59.53 & 54.27 & 69.01 \\
% \midrule
% Dataset 25\% (VLM-3R) & 54.0 & 41.7 & 47.3 \\
% Dataset 25\% (Ours) & \textbf{58.8} & \textbf{53.0} &  \textbf{70.9} \\
% % Dataset 25\% (Ours) & 58.81 & 53.05 &  70.91 \\
% \midrule
% Dataset 10\% (VLM-3R) & 49.1 & 37.3 & 47.0 \\
% Dataset 10\% (Ours) & \textbf{54.5} & \textbf{51.5} &  \textbf{70.0} \\
% \bottomrule
% \end{tabular}%
% }
% \end{table}

\begin{table}[htbp]
\centering
\small
\caption{Ablation study across varying training data fractions. We compare our method against VLM-3R. \textbf{Bold} denotes better results.}
\label{tab:data_frac}
\setlength{\tabcolsep}{6pt} % 增加列间距，不要用 0.8mm
\resizebox{\linewidth}{!}{
\begin{tabular}{l l c c c} % l l c c c 结构
\toprule
\multicolumn{1}{c}{\textbf{Data Fraction}} & \multicolumn{1}{c}{\textbf{Method}} & \textbf{Average} & \textbf{Abs. Dist} & \textbf{Rel. Dir} \\
\midrule

% 100% Block
\multirow{2}{*}{100\% Dataset} & VLM-3R & 60.9 & 49.4 & \textbf{80.5} \\
 & {Ours} & \textbf{61.0} & \textbf{55.0} & 69.8 \\
\midrule

% 50% Block
\multirow{2}{*}{50\% Dataset} & VLM-3R & 55.5 & 43.7 & 52.5 \\
 & {Ours} & \textbf{59.5} & \textbf{54.2} & \textbf{69.0} \\
\midrule

% 25% Block
\multirow{2}{*}{25\% Dataset} & VLM-3R & 54.0 & 41.7 & 47.3 \\
 & {Ours} & \textbf{58.8} & \textbf{53.0} & \textbf{70.9} \\
\midrule

% 10% Block
\multirow{2}{*}{10\% Dataset} & VLM-3R & 49.1 & 37.3 & 47.0 \\
 & {Ours} & \textbf{54.5} & \textbf{51.5} & \textbf{70.0} \\

\bottomrule
\end{tabular}
}
\end{table}

\vspace{-15pt}
\paragraph{Ablation on Metric-CogMap and Cog-CoT}

We assess the contributions of the \emph{Metric-CogMap} and \emph{Cog-CoT} components in Tab.~\ref{tab:ablation}. All experiments in this table are trained using 25\% of the full dataset.

Experiment (1) represents our complete model, which uses a predicted \emph{Metric-CogMap} constructed via our proposed pipeline and applies \emph{Cog-CoT} for stepwise reasoning before answer generation.
Experiment (2) disables the \emph{Cog-CoT} module while retaining the predicted \emph{Metric-CogMap}, thus removing explicit cognitive reasoning and relying solely on spatial metric cues.
And, experiment (3) removes both the \emph{Metric-CogMap} and \emph{Cog-CoT}, which is the same as the VLM-3R baseline setting. 
This confirms that both components play complementary and essential roles in enhancing spatial reasoning, with \emph{Cog-CoT} providing the necessary cognitive reasoning framework and \emph{Metric-CogMap} offering a structured spatial representation. Their combined use allows our model to outperform existing methods, showcasing the importance of integrating explicit spatial and stepwise cognitive inference. 

Experiment (4) replaces the learned \emph{Metric-CogMap} with a grid-based cognitive map used in VSI-Bench~\cite{vsibench} and MindCube~\cite{mindcube}, leading to further performance degradation. This degradation indicates that coarse discretized spatial representations are insufficient for capturing continuous metric structure and fine-grained geometric relationships.
Finally, experiment (5) uses ground-truth annotations to construct a perfect \emph{Metric-CogMap}, providing an upper-bound estimate of our framework’s performance. This result demonstrates the potential of our method when supplied with ideal spatial priors, while also underscoring the limitations of current 2D and 3D detection/segmentation algorithms—particularly in tasks like \emph{Rel.Dir} and \emph{Rel.Dist}, as shown in Tab.~\ref{tab:vsibench}. This points to a promising research direction: bridging the gap between state-of-the-art visual perception and ground-truth-level Metric-CogMap construction to further enhance the Map2Thought framework.

% \begin{table}[htbp]
% \centering
% \small
% \caption{Ablation study on model components. All models are trained using 25\% dataset.} 
% \label{tab:ablation}
% \setlength{\tabcolsep}{0.8 mm}{
% \begin{tabular}{lccc}
% \toprule
% \multicolumn{1}{c}{\multirow{1}{*}{Method}} & \rotatebox{50}{Average} & \rotatebox{50}{Abs. Dist} & \rotatebox{50}{Rel. Dir} \\
% \midrule

% % GT Metric-CogMap & 73.75 & 81.31 & 86.11 \\
% % Full Model & 58.81 & 53.05 & 70.91 \\
% % Metric-CogMap, w/o Cog-CoT & 54.08 & 41.97 & 45.89 \\
% % Grid CogMap, w/o Cog-CoT & 49.70  & 19.40 & 46.42 \\
% % w/o CogMap, w/o Cog-CoT & 53.98 & 41.71 & 47.38 \\

% (1) Pred. \emph{Metric-CogMap}, \emph{Cog-CoT} & 58.8 & 53.0 & 70.9 \\
% (2) Pred. \emph{Metric-CogMap}, w/o \emph{Cog-CoT} & 54.0 & 41.9 & 45.9 \\
% (3) w/o \emph{Metric CogMap}, w/o \emph{Cog-CoT} & 54.0 & 41.7 & 47.4 \\
% (4) Grid \emph{CogMap}, w/o \emph{Cog-CoT} & 49.7  & 19.4 & 46.4 \\
% (5) GT. \emph{Metric-CogMap}, \emph{Cog-CoT}  & 73.7 & 81.3 & 86.1 \\

% \bottomrule
% \end{tabular}
% }
% \end{table}

\begin{table}[htbp] % [t] 通常放在每一页的顶部
\centering
% 引入 resizebox 需要 graphicx 包
% \usepackage{graphicx}
\caption{Ablation study. We compare different map inputs and the effect of Cog-CoT.}
\label{tab:ablation}
\resizebox{\linewidth}{!}{ % 核心：强制缩放到栏宽
\begin{tabular}{l | c c | c c c}
\toprule
\multicolumn{1}{c|}{\multirow{2}{*}{\textbf{Method}}} & \multicolumn{2}{c|}{\textbf{Components}} & \multicolumn{3}{c}{\textbf{Metrics}} \\
 & \small{Map Type} & \small{Cog-CoT} & Avg & Abs.Dist & Rel.Dir \\
\midrule
% Row 1: Full Model (Highlight)
(1) {Ours} & {Pred} & \checkmark & \cellcolor{yellow!30}\textbf{58.8} & \cellcolor{yellow!30}\textbf{53.0} & \cellcolor{yellow!30}\textbf{70.9} \\

% Row 2: Metric Map only
(2) Variant & Pred & - & 54.0 & 41.9 & 45.9 \\

% Row 3: Baseline
(3) Baseline & None & - & 54.0 & 41.7 & 47.4 \\

% Row 4: Grid
(4) Variant & Grid & - & 49.7 & 19.4 & 46.4 \\

\midrule
% Row 5: GT
(5) Upper Bound & GT & \checkmark & \cellcolor{orange!30}73.7 & \cellcolor{orange!30}81.3 & \cellcolor{orange!30}86.1 \\
\bottomrule
\end{tabular}
}
\end{table}

\vspace{-10pt}
\section{Conclusion}
\label{sec:conclusion}

In this work, we introduced Map2Thought, a framework that enables explicit and interpretable 3D spatial reasoning for vision-language models. By integrating the \emph{Metric-CogMap} with \emph{Cog-CoT} for structured geometric reasoning, our method supports transparent and verifiable inference. We also proposed a video-to-map extraction pipeline that bridges 2D perception and 3D geometry. Extensive experiments demonstrate that Map2Thought achieves strong data efficiency and superior generalization, advancing the goal of reliable, explainable spatial reasoning in 3D vision-language understanding. While our method shows robust performance, we observed that its effectiveness is bounded by the quality of the reconstructed \emph{Metric-CogMap}. Future work will explore more accurate and robust map construction to further enhance spatial reasoning reliability.

{
    \small
    \bibliographystyle{ieeenat_fullname}
    \bibliography{main}

@String(CVPR= {IEEE Conf. Comput. Vis. Pattern Recog.})

@String(ICCV= {Int. Conf. Comput. Vis.})

@String(ECCV= {Eur. Conf. Comput. Vis.})

@String(ICLR = {Int. Conf. Learn. Represent.})

@String(CVPR  = {CVPR})

@String(ICCV  = {ICCV})

@String(ECCV  = {ECCV})

@String(ICLR  = {ICLR})

@inproceedings{dai2017scannet,
 author = {Angela Dai and
Angel X. Chang and
Manolis Savva and
Maciej Halber and
Thomas A. Funkhouser and
Matthias Nie{\ss}ner},
 booktitle = {CVPR},
 title = {ScanNet: Richly-Annotated 3D Reconstructions of Indoor Scenes},
 year = {2017}
}

@inproceedings{azuma2022scanqa,
  title={Scanqa: 3d question answering for spatial scene understanding},
  author={Azuma, Daichi and Miyanishi, Taiki and Kurita, Shuhei and Kawanabe, Motoaki},
  booktitle={proceedings of the IEEE/CVF conference on computer vision and pattern recognition},
  pages={19129--19139},
  year={2022}
}

@article{vsibench,
 author = {Jihan Yang and
Shusheng Yang and
Anjali W. Gupta and
Rilyn Han and
Li Fei{-}Fei and
Saining Xie},
 journal = {arXiv:2412.14171},
 title = {Thinking in Space: How Multimodal Large Language Models See, Remember,
and Recall Spaces},
 year = {2024}
}

@article{qwen2_5_vl,
 author = {Bai, Shuai and Chen, Keqin and Liu, Xuejing and Wang, Jialin and Ge, Wenbin and Song, Sibo and Dang, Kai and Wang, Peng and Wang, Shijie and Tang, Jun and others},
 journal = {arXiv:2502.13923},
 title = {Qwen2.5-VL Technical Report},
 year = {2025}
}

@article{gpt4o,
 author={Hurst, Aaron and Lerer, Adam and Goucher, Adam P and Perelman, Adam and Ramesh, Aditya and Clark, Aidan and Ostrow, AJ and Welihinda, Akila and Hayes, Alan and Radford, Alec and others},
 journal = {arXiv:2410.21276},
 title = {GPT-4o System Card},
 year = {2024}
}

@article{llava_onevision,
 author = {Bo Li and
Yuanhan Zhang and
Dong Guo and
Renrui Zhang and
Feng Li and
Hao Zhang and
Kaichen Zhang and
Yanwei Li and
Ziwei Liu and
Chunyuan Li},
 journal = {arXiv:2408.03326},
 title = {LLaVA-OneVision: Easy Visual Task Transfer},
 year = {2024}
}

@article{gemini_pro,
 author={Team, Gemini and Georgiev, Petko and Lei, Ving Ian and Burnell, Ryan and Bai, Libin and Gulati, Anmol and Tanzer, Garrett and Vincent, Damien and Pan, Zhufeng and Wang, Shibo and others},
 journal = {arXiv:2403.05530},
 title = {Gemini 1.5: Unlocking multimodal understanding across millions of
tokens of context},
 year = {2024}
}

@inproceedings{vila,
 author = {Ji Lin and
Hongxu Yin and
Wei Ping and
Pavlo Molchanov and
Mohammad Shoeybi and
Song Han},
 booktitle = {CVPR},
 title = {{VILA:} On Pre-training for Visual Language Models},
 year = {2024}
}

@article{llava3d,
 author = {Chenming Zhu and
Tai Wang and
Wenwei Zhang and
Jiangmiao Pang and
Xihui Liu},
 journal = {arXiv:2409.18125},
 title = {LLaVA-3D: {A} Simple yet Effective Pathway to Empowering LMMs with
3D-awareness},
 year = {2024}
}

@inproceedings{video3dllm,
 author = {Duo Zheng and
Shijia Huang and
Liwei Wang},
 booktitle = {CVPR},
 title = {Video-3D {LLM:} Learning Position-Aware Video Representation for 3D
Scene Understanding},
 year = {2025}
}

@article{gpt4scene,
 author = {Zhangyang Qi and
Zhixiong Zhang and
Ye Fang and
Jiaqi Wang and
Hengshuang Zhao},
 journal = {arXiv:2501.01428},
 title = {GPT4Scene: Understand 3D Scenes from Videos with Vision-Language Models},
 year = {2025}
}

@inproceedings{chen2022vil3drel,
 author = {Shizhe Chen and
Pierre{-}Louis Guhur and
Makarand Tapaswi and
Cordelia Schmid and
Ivan Laptev},
 booktitle = {NeurIPS},
 title = {Language Conditioned Spatial Relation Reasoning for 3D Object Grounding},
 year = {2022}
}

@inproceedings{leo,
 author = {Jiangyong Huang and
Silong Yong and
Xiaojian Ma and
Xiongkun Linghu and
Puhao Li and
Yan Wang and
Qing Li and
Song{-}Chun Zhu and
Baoxiong Jia and
Siyuan Huang},
 booktitle = {ICML},
 title = {An Embodied Generalist Agent in 3D World},
 year = {2024}
}

@inproceedings{wang2025vggt,
  title={{VGGT}: Visual Geometry Grounded Transformer},
  author={Wang, Jianyuan and Chen, Minghao and Karaev, Nikita and Vedaldi, Andrea and Rupprecht, Christian and Novotny, David},
  booktitle={Proceedings of the IEEE/CVF Conference on Computer Vision and Pattern Recognition},
  year={2025}
}

@inproceedings{scannetpp,
 author = {Chandan Yeshwanth and
Yueh{-}Cheng Liu and
Matthias Nie{\ss}ner and
Angela Dai},
 booktitle = {ICCV},
 title = {ScanNet++: {A} High-Fidelity Dataset of 3D Indoor Scenes},
 year = {2023}
}

@inproceedings{arkitscenes,
 author = {Afshin Dehghan and
Gilad Baruch and
Zhuoyuan Chen and
Yuri Feigin and
Peter Fu and
Thomas Gebauer and
Daniel Kurz and
Tal Dimry and
Brandon Joffe and
Arik Schwartz and
Elad Shulman},
 booktitle = {NeurIPS},
 title = {ARKitScenes: {A} Diverse Real-World Dataset For 3D Indoor Scene Understanding
Using Mobile {RGB-D} Data},
 year = {2021}
}

@misc{wang2025ross3d,
 author = {Haochen Wang and Yucheng Zhao and Tiancai Wang and Haoqiang Fan and Xiangyu Zhang and Zhaoxiang Zhang},
 journal = {arXiv:2504.01901},
 title = {Ross3D: Reconstructive Visual Instruction Tuning with 3D-Awareness},
 year = {2025}
}

@article{fan2025vlm,
  title={{VLM-3R}: Vision-Language Models Augmented with Instruction-Aligned 3D Reconstruction},
  author={Fan, Zhiwen and Zhang, Jian and Li, Renjie and Zhang, Junge and Chen, Runjin and Hu, Hezhen and Wang, Kevin and Qu, Huaizhi and Wang, Dilin and Yan, Zhicheng and others},
  journal={arXiv preprint arXiv:2505.20279},
  year={2025}
}

@inproceedings{radford2021clip,
  title={Learning transferable visual models from natural language supervision},
  author={Radford, Alec and Kim, Jong Wook and Hallacy, Chris and Ramesh, Aditya and Goh, Gabriel and Agarwal, Sandhini and Sastry, Girish and Askell, Amanda and Mishkin, Pamela and Clark, Jack and others},
  booktitle=ICML,
  year={2021},
}

@inproceedings{jia2021align,
  title={Scaling up visual and vision-language representation learning with noisy text supervision},
  author={Jia, Chao and Yang, Yinfei and Xia, Ye and Chen, Yi-Ting and Parekh, Zarana and Pham, Hieu and Le, Quoc and Sung, Yun-Hsuan and Li, Zhen and Duerig, Tom},
  booktitle=ICML,
  year={2021},
}

@article{alayrac2022flamingo,
  title={Flamingo: a visual language model for few-shot learning},
  author={Alayrac, Jean-Baptiste and Donahue, Jeff and Luc, Pauline and Miech, Antoine and Barr, Iain and Hasson, Yana and Lenc, Karel and Mensch, Arthur and Millican, Katherine and Reynolds, Malcolm and others},
  booktitle=NeurIPS,
  year={2022}
}

@inproceedings{hong20233d,
  title={3d-llm: Injecting the 3d world into large language models},
  author={Hong, Yining and Zhen, Haoyu and Chen, Peihao and Zheng, Shuhong and Du, Yilun and Chen, Zhenfang and Gan, Chuang},
  booktitle={NeurIPS},
  year={2023}
}

@article{ye20223d,
  title={3D question answering},
  author={Ye, Shuquan and Chen, Dongdong and Han, Songfang and Liao, Jing},
  journal={IEEE Transactions on Visualization and Computer Graphics},
  volume={30},
  number={3},
  pages={1772--1786},
  year={2022},
  publisher={IEEE}
}

@article{chat3d,
  title={Chat-3d: Data-efficiently tuning large language model for universal dialogue of 3d scenes},
  author={Wang, Zehan and Huang, Haifeng and Zhao, Yang and Zhang, Ziang and Zhao, Zhou},
  journal={arXiv preprint arXiv:2308.08769},
  year={2023}
}

@inproceedings{deng20253d,
  title={3d-llava: Towards generalist 3d lmms with omni superpoint transformer},
  author={Deng, Jiajun and He, Tianyu and Jiang, Li and Wang, Tianyu and Dayoub, Feras and Reid, Ian},
  booktitle={Proceedings of the Computer Vision and Pattern Recognition Conference},
  pages={3772--3782},
  year={2025}
}

@inproceedings{chen2024ll3da,
  title={Ll3da: Visual interactive instruction tuning for omni-3d understanding reasoning and planning},
  author={Chen, Sijin and Chen, Xin and Zhang, Chi and Li, Mingsheng and Yu, Gang and Fei, Hao and Zhu, Hongyuan and Fan, Jiayuan and Chen, Tao},
  booktitle={Proceedings of the IEEE/CVF conference on computer vision and pattern recognition},
  pages={26428--26438},
  year={2024}
}

@inproceedings{li20243dmit,
  title={3dmit: 3d multi-modal instruction tuning for scene understanding},
  author={Li, Zeju and Zhang, Chao and Wang, Xiaoyan and Ren, Ruilong and Xu, Yifan and Ma, Ruifei and Liu, Xiangde and Wei, Rong},
  booktitle={2024 IEEE International Conference on Multimedia and Expo Workshops (ICMEW)},
  pages={1--5},
  year={2024},
  organization={IEEE}
}

@article{huang2025mllms,
  title={MLLMs Need 3D-Aware Representation Supervision for Scene Understanding},
  author={Huang, Xiaohu and Wu, Jingjing and Xie, Qunyi and Han, Kai},
  journal={arXiv preprint arXiv:2506.01946},
  year={2025}
}

@InProceedings{cut3r,
    author    = {Wang, Qianqian and Zhang, Yifei and Holynski, Aleksander and Efros, Alexei A. and Kanazawa, Angjoo},
    title     = {Continuous 3D Perception Model with Persistent State},
    booktitle = {Proceedings of the IEEE/CVF Conference on Computer Vision and Pattern Recognition (CVPR)},
    month     = {June},
    year      = {2025},
    pages     = {10510-10522}
}

@article{hu2022lora,
  title={Lora: Low-rank adaptation of large language models.},
  author={Hu, Edward J and Shen, Yelong and Wallis, Phillip and Allen-Zhu, Zeyuan and Li, Yuanzhi and Wang, Shean and Wang, Lu and Chen, Weizhu and others},
  journal={ICLR},
  volume={1},
  number={2},
  pages={3},
  year={2022}
}

@inproceedings{mindcube,
  title={Spatial mental modeling from limited views},
  author={Yin, Baiqiao and Wang, Qineng and Zhang, Pingyue and Zhang, Jianshu and Wang, Kangrui and Wang, Zihan and Zhang, Jieyu and Chandrasegaran, Keshigeyan and Liu, Han and Krishna, Ranjay and others},
  booktitle={Structural Priors for Vision Workshop at ICCV'25},
  year={2025}
}

@misc{wu2025spatialmllmboostingmllmcapabilities,
      title={Spatial-MLLM: Boosting MLLM Capabilities in Visual-based Spatial Intelligence}, 
      author={Diankun Wu and Fangfu Liu and Yi-Hsin Hung and Yueqi Duan},
      year={2025},
      eprint={2505.23747},
      archivePrefix={arXiv},
      primaryClass={cs.CV},
      url={https://arxiv.org/abs/2505.23747}, 
}

@article{vgllm,
  title={Learning from Videos for 3D World: Enhancing MLLMs with 3D Vision Geometry Priors},
  author={Zheng, Duo and Huang, Shijia and Li, Yanyang and Wang, Liwei},
  journal={arXiv preprint arXiv:2505.24625},
  year={2025}
}

@article{chen2024internvl2,
  title={How far are we to gpt-4v? closing the gap to commercial multimodal models with open-source suites},
  author={Chen, Zhe and Wang, Weiyun and Tian, Hao and Ye, Shenglong and Gao, Zhangwei and Cui, Erfei and Tong, Wenwen and Hu, Kongzhi and Luo, Jiapeng and Ma, Zheng and others},
  journal={arXiv preprint arXiv:2404.16821},
  year={2024}
}

@article{zhang2024longva,
  title={Long context transfer from language to vision},
  author={Zhang, Peiyuan and Zhang, Kaichen and Li, Bo and Zeng, Guangtao and Yang, Jingkang and Zhang, Yuanhan and Wang, Ziyue and Tan, Haoran and Li, Chunyuan and Liu, Ziwei},
  journal={arXiv preprint arXiv:2406.16852},
  year={2024}
}

@misc{zhang2024llavanextvideo,
  title={{LLaVA-NeXT}: A Strong Zero-shot Video Understanding Model},
  url={https://llava-vl.github.io/blog/2024-04-30-llava-next-video/},
  author={Zhang, Yuanhan and Li, Bo and Liu, haotian and Lee, Yong jae and Gui, Liangke and Fu, Di and Feng, Jiashi and Liu, Ziwei and Li, Chunyuan},
  year={2024}
}

@article{touvron2023llama2,
  title={Llama 2: Open foundation and fine-tuned chat models},
  author={Touvron, Hugo and Martin, Louis and Stone, Kevin and Albert, Peter and Almahairi, Amjad and Babaei, Yasmine and Bashlykov, Nikolay and Batra, Soumya and Bhargava, Prajjwal and Bhosale, Shruti and others},
  journal={arXiv preprint arXiv:2307.09288},
  year={2023}
}

@inproceedings{li2023blip2,
  title={Blip-2: Bootstrapping language-image pre-training with frozen image encoders and large language models},
  author={Li, Junnan and Li, Dongxu and Savarese, Silvio and Hoi, Steven},
  booktitle={ICML},
  year={2023}
}

@inproceedings{zhou2022detecting,
  title={Detecting Twenty-thousand Classes using Image-level Supervision},
  author={Zhou, Xingyi and Girdhar, Rohit and Joulin, Armand and Kr{\"a}henb{\"u}hl, Philipp and Misra, Ishan},
  booktitle={ECCV},
  year={2022}
}

@inproceedings{liu2024grounding,
  title = {{Grounding DINO}: Marrying DINO with Grounded Pre-training for Open-Set Object Detection},
  author = {Liu, Shilong and Zeng, Zhaoyang and Ren, Tianhe and Li, Feng and Zhang, Hao and Yang, Jie and Jiang, Qing and Li, Chunyuan and Yang, Jianwei and Su, Hang and Zhu, Jun and Zhang, Lei},
  booktitle={ECCV},
  pages = {38–55},
  numpages = {18},
  year={2024}
}

@InProceedings{Jang_2025_CVPR,
    author    = {Jang, Youngkyoon and P\'erez-Pellitero, Eduardo},
    title     = {{CoMapGS}: Covisibility Map-based Gaussian Splatting for Sparse Novel View Synthesis},
    booktitle = {Proceedings of the Computer Vision and Pattern Recognition Conference (CVPR)},
    month     = {June},
    year      = {2025},
    pages     = {26779-26788}
}

@misc{Inst3D-LMM,
    title={Inst3D-LMM: Instance-Aware 3D Scene Understanding with Multi-modal Instruction Tuning}, 
    author={Hanxun Yu and Wentong Li and Song Wang and Junbo Chen and Jianke Zhu},
    year={2025},
    eprint={2503.00513},
    archivePrefix={arXiv},
    primaryClass={cs.CV},
    url={https://arxiv.org/abs/2503.00513}, 
}

@misc{wang2025pi3,
      title={$\pi^3$: Scalable Permutation-Equivariant Visual Geometry Learning}, 
      author={Yifan Wang and Jianjun Zhou and Haoyi Zhu and Wenzheng Chang and Yang Zhou and Zizun Li and Junyi Chen and Jiangmiao Pang and Chunhua Shen and Tong He},
      year={2025},
      eprint={2507.13347},
      archivePrefix={arXiv},
      primaryClass={cs.CV},
      url={https://arxiv.org/abs/2507.13347}, 
}

@InProceedings{ravi2025sam2,
    title={{SAM 2}: Segment Anything in Images and Videos},
    author={Ravi, Nikhila and Gabeur, Valentin and Hu, Yuan-Ting and Hu, Ronghang and Ryali, Chaitanya and Ma, Tengyu and Khedr, Haitham and R{\"a}dle, Roman and Rolland, Chloe and Gustafson, Laura and Mintun, Eric and Pan, Junting and Alwala, Kalyan Vasudev and Carion, Nicolas and Wu, Chao-Yuan and Girshick, Ross and Doll{\'a}r, Piotr and Feichtenhofer, Christoph},
    booktitle={ICLR},
    year      = {2025}
}

@InProceedings{cao2024cognav,
  title={CogNav: Cognitive Process Modeling for Object Goal Navigation with LLMs},
  author={Cao, Yihan and Zhang, Jiazhao and Yu, Zhinan and Liu, Shuzhen and Qin, Zheng and Zou, Qin and Du, Bo and Xu, Kai},
  booktitle={ICCV},
  year={2025}
}

@InProceedings{zhang2025spatial,
  title={Spatial Understanding from Videos: Structured Prompts Meet Simulation Data},
  author={Zhang, Haoyu and Liu, Meng and Li, Zaijing and Wen, Haokun and Guan, Weili and Wang, Yaowei and Nie, Liqiang},
  booktitle={NeurIPS},
  year={2025}
}

@misc{wang2025moge2accuratemonoculargeometry,
      title={{MoGe-2}: Accurate Monocular Geometry with Metric Scale and Sharp Details}, 
      author={Ruicheng Wang and Sicheng Xu and Yue Dong and Yu Deng and Jianfeng Xiang and Zelong Lv and Guangzhong Sun and Xin Tong and Jiaolong Yang},
      year={2025},
      eprint={2507.02546},
      archivePrefix={arXiv},
      primaryClass={cs.CV},
      url={https://arxiv.org/abs/2507.02546}, 
}

@misc{vlm3r,
      title={{VLM-3R}: Vision-Language Models Augmented with Instruction-Aligned 3D Reconstruction}, 
      author={Zhiwen Fan and Jian Zhang and Renjie Li and Junge Zhang and Runjin Chen and Hezhen Hu and Kevin Wang and Huaizhi Qu and Dilin Wang and Zhicheng Yan and Hongyu Xu and Justin Theiss and Tianlong Chen and Jiachen Li and Zhengzhong Tu and Zhangyang Wang and Rakesh Ranjan},
      year={2025},
      eprint={2505.20279},
      archivePrefix={arXiv},
      primaryClass={cs.CV},
      url={https://arxiv.org/abs/2505.20279}, 
}

@inproceedings{chen2020scanrefer,
  title={Scanrefer: 3d object localization in rgb-d scans using natural language},
  author={Chen, Dave Zhenyu and Chang, Angel X and Nie{\ss}ner, Matthias},
  booktitle={European conference on computer vision},
  pages={202--221},
  year={2020},
  organization={Springer}
}

@inproceedings{chen2021scan2cap,
  title={Scan2cap: Context-aware dense captioning in rgb-d scans},
  author={Chen, Zhenyu and Gholami, Ali and Nie{\ss}ner, Matthias and Chang, Angel X},
  booktitle={Proceedings of the IEEE/CVF conference on computer vision and pattern recognition},
  pages={3193--3203},
  year={2021}
}

@inproceedings{chen2022d3net,
  title={D 3 net: A unified speaker-listener architecture for 3d dense captioning and visual grounding},
  author={Chen, Dave Zhenyu and Wu, Qirui and Nie{\ss}ner, Matthias and Chang, Angel X},
  booktitle={European Conference on Computer Vision},
  pages={487--505},
  year={2022},
  organization={Springer}
}

@inproceedings{chen2023unit3d,
  title={Unit3d: A unified transformer for 3d dense captioning and visual grounding},
  author={Chen, Zhenyu and Hu, Ronghang and Chen, Xinlei and Nie{\ss}ner, Matthias and Chang, Angel X},
  booktitle={Proceedings of the IEEE/CVF international conference on computer vision},
  pages={18109--18119},
  year={2023}
}

@article{ma2024llms,
  title={When llms step into the 3d world: A survey and meta-analysis of 3d tasks via multi-modal large language models},
  author={Ma, Xianzheng and Smart, Brandon and Bhalgat, Yash and Chen, Shuai and Li, Xinghui and Ding, Jian and Gu, Jindong and Chen, Dave Zhenyu and Peng, Songyou and Bian, Jia-Wang and others},
  journal={arXiv preprint arXiv:2405.10255},
  year={2024}
}

@article{dwedari2023generating,
  title={Generating context-aware natural answers for questions in 3D scenes},
  author={Dwedari, Mohammed Munzer and Niessner, Matthias and Chen, Dave Zhenyu},
  journal={arXiv preprint arXiv:2310.19516},
  year={2023}
}

@article{li2025does,
  title={Does Your 3D Encoder Really Work? When Pretrain-SFT from 2D VLMs Meets 3D VLMs},
  author={Li, Haoyuan and Zhou, Yanpeng and Gao, Yufei and Tang, Tao and Han, Jianhua and Yuan, Yujie and Chen, Dave Zhenyu and Bian, Jiawang and Xu, Hang and Liang, Xiaodan},
  journal={arXiv preprint arXiv:2506.05318},
  year={2025}
}

@article{xu2025uniugg,
  title={Uniugg: Unified 3d understanding and generation via geometric-semantic encoding},
  author={Xu, Yueming and Zhang, Jiahui and Huang, Ze and Chen, Yurui and Zhou, Yanpeng and Chen, Zhenyu and Yuan, Yu-Jie and Xia, Pengxiang and Huang, Guowei and Cai, Xinyue and others},
  journal={arXiv preprint arXiv:2508.11952},
  year={2025}
}
}

\clearpage
\newpage
\appendix

\setcounter{figure}{0}
\setcounter{table}{0}
\renewcommand{\thefigure}{S\arabic{figure}}
\renewcommand{\thetable}{S\arabic{table}}

\setcounter{page}{1}
\maketitlesupplementary

\section*{Appendix Overview}
% \textcolor{red}{(Xiangjun, Please revise the concise explanations (blue text) reflecting the contents in the appendix.)} 
\textcolor{black}{This supplementary document provides additional implementation details and results that support and extend the main paper.} It is organized as follows: 
\begin{itemize}
    \item \textbf{Supp: Additional \emph{Metric-CogMap} examples for evaluation}. Qualitative visualizations of constructed \emph{Metric-CogMap}s across datasets.
    \item \textbf{Supp: \emph{Metric-CogMap} construction for training}. \textcolor{black}{Details of how \emph{Metric-CogMap} annotations are generated for Map2Thought.}
    % \item \textbf{Supp: Training dataset construction}. \textcolor{black}{Details of how training scenes and \emph{Metric-CogMap} annotations are generated for Map2Thought.}
    \item \textbf{Supp: \emph{Cog-CoT} construction and Complete QA.} \textcolor{black}{Details of \emph{Cog-CoT} construction. Also, complete examples of questions, maps, and corresponding reasoning traces.}
    % \item \textbf{Supp: Complete QA, \emph{Metric-CogMap}, and \emph{Cog-CoT}}. \textcolor{black}{Complete examples of questions, maps, and corresponding reasoning traces.}
    \item \textbf{Supp: Zero-shot Qualitative Demonstrations}. \textcolor{black}{Qualitative examples showing how \emph{Metric-CogMap} and \emph{Cog-CoT} improve zero-shot spatial reasoning with commercial VLMs.}
    % \item \textbf{Supp: Extended Benchmarks}. \textcolor{blue}{Evaluation results on additional datasets beyond VSI-Bench.}
\end{itemize}

\section{Additional \emph{Metric-CogMap} examples for evaluation}

\begin{figure*}[htpb]
	\centering
	\includegraphics[width=1.0 \linewidth]{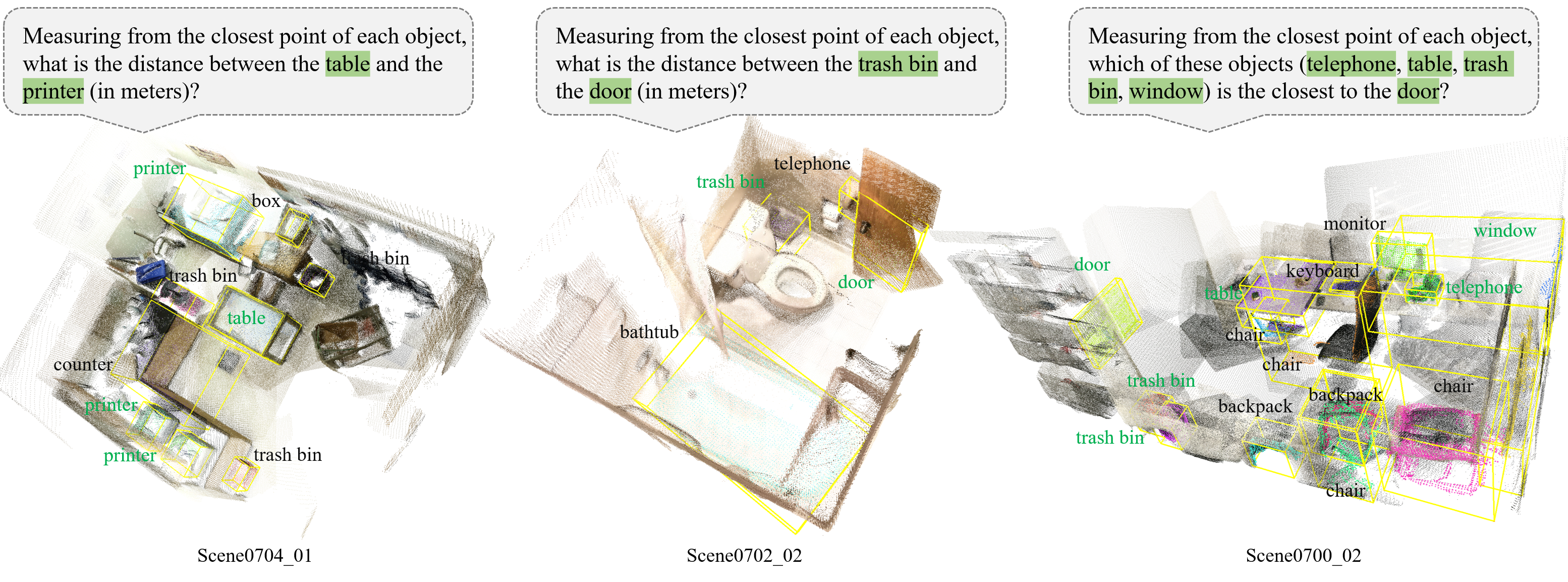} \\
    \includegraphics[width=1.0 \linewidth]{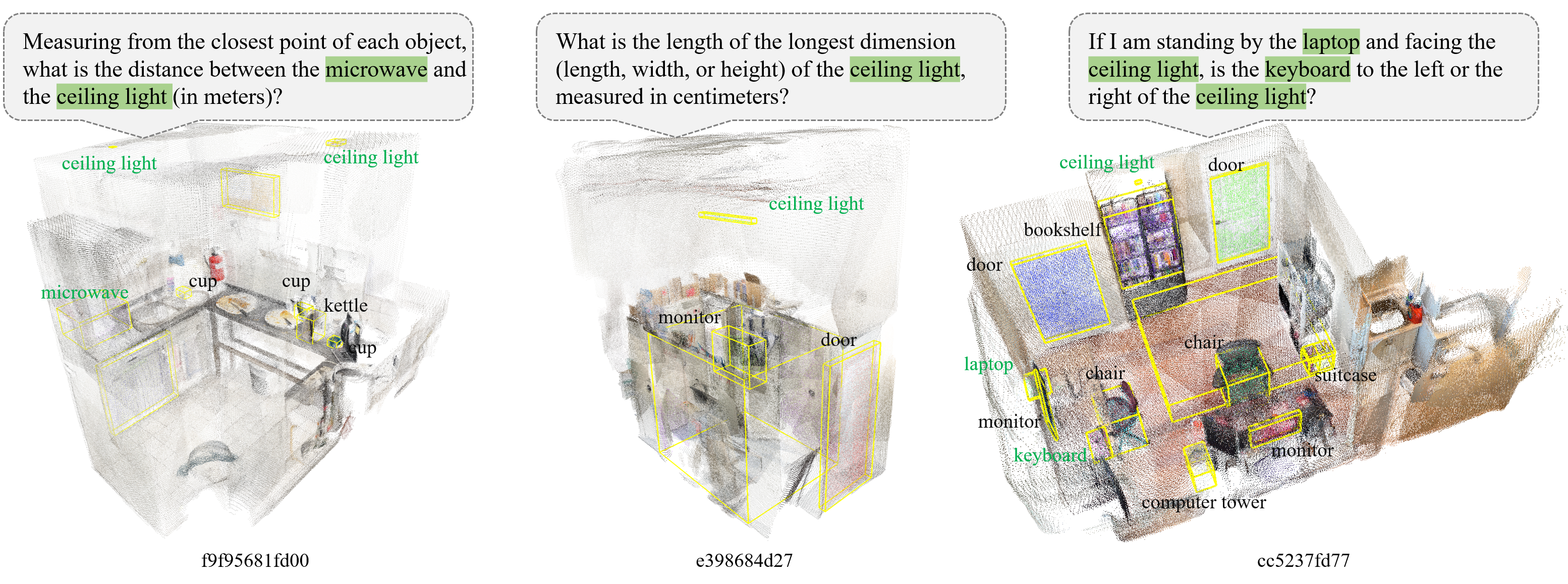} \\
    \includegraphics[width=1.0 \linewidth]{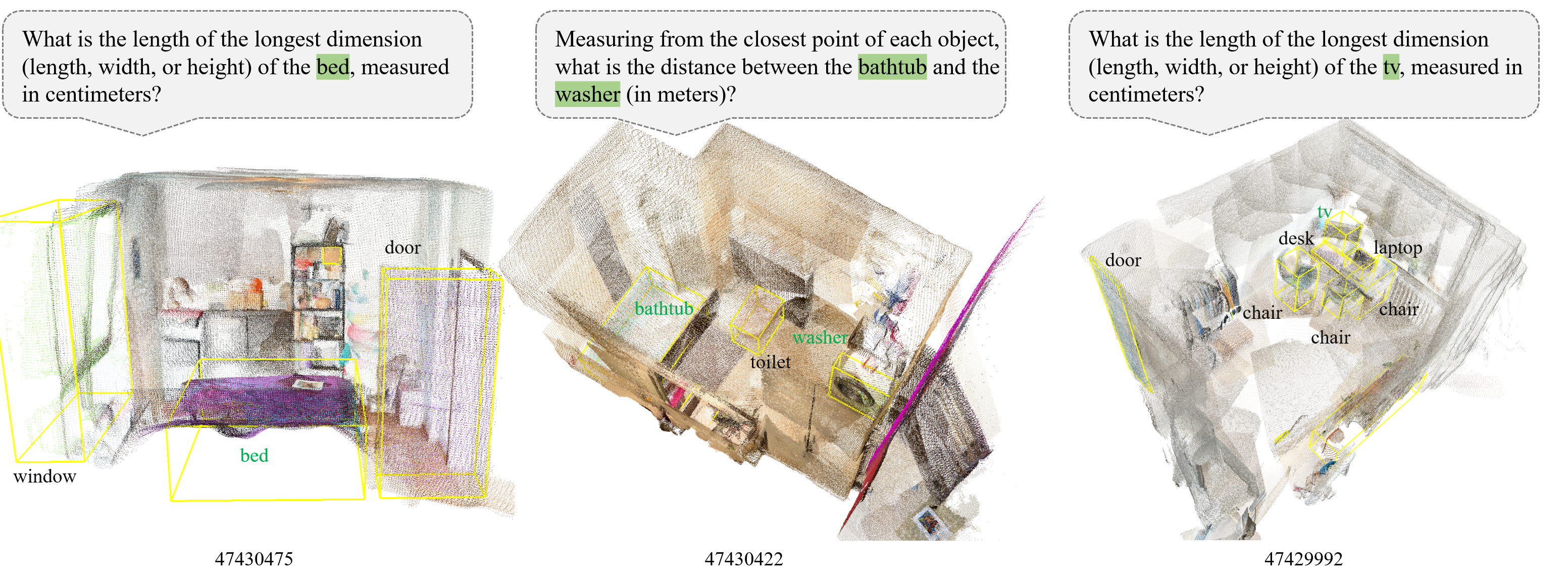} \\
        \caption{\textbf{Examples of evaluation \emph{Metric-CogMap} outputs on the VSI-Bench~\citep{vsibench}.} Each row showcases a representative scene from one of the three datasets used in VSI-Bench: (top) ScanNet, (middle) ScanNet++, and (bottom) ARKitScenes. Above each scene visualization, we display a sample query with highlighted object mentions (green background). Within the corresponding 3D visualization, the mentioned objects are labeled in green text and enclosed in yellow axis-aligned bounding boxes, as produced by the \emph{Metric-CogMap} construction pipeline. Additionally, objects mentioned in other queries from the same scene are shown in gray text. This reflects the unified nature of the \emph{Metric-CogMap}, which integrates all detected and reconstructed objects relevant to multiple queries within a single scene-level map.}
        \label{fig:metric_cogmap_examples}
\end{figure*}

While the main paper provides an overview of our \emph{Metric-CogMap} construction pipeline, space constraints limited the inclusion of more visual examples. To address this, we include additional visualizations as shown in Fig.~\ref{fig:metric_cogmap_examples}, showcasing scene-level \emph{Metric-CogMaps} produced by our evaluation-time pipeline across the three datasets in VSI-Bench~\citep{vsibench}. These examples demonstrate how our pipeline identifies and integrates query-relevant objects into a unified, metrically accurate representation, which supports downstream spatial reasoning tasks across multiple queries within each scene. Note that the construction pipeline described in the main manuscript (Sec.\textcolor{red}{3.3}) is specific to the evaluation phase; a separate, ground-truth annotation-based procedure is used during training, independent of the evaluation-time construction process, as detailed in Sec.~\ref{Supp:dataset_construction}.

% \section{Training dataset construction}
\section{\emph{Metric-CogMap} construction for training}

\label{Supp:dataset_construction}

We build our training dataset upon VLM-3R~\cite{vlm3r}, which provides instructional question-answer pairs from ScanNet~\cite{dai2017scannet}, ScanNet++~\cite{scannetpp}, and ARKitScenes~\cite{arkitscenes}. As noted earlier, while Sec.\textcolor{red}{3.3} in the main manuscript outlines the \emph{Metric-CogMap} construction process for evaluation, this section details the complementary pipeline used for the training dataset construction. 
% Specifically, Sec.\textcolor{red}{3.3}(Sec.\textcolor{red}{3.3}) focuses on the inference-time map generation process for benchmark evaluation, whereas here we elaborate on the complete data processing workflow for training, including metadata extraction from raw 3D data, scene boundary computation, grid-based map discretization, object category selection, and chain-of-thought supervision generation.
To facilitate explicit spatial computation in Cog-CoT inference, we augment each training scene with a ground-truth \emph{Metric-CogMap}. The construction process involves the following steps:

\paragraph{Metadata extraction.} 

To construct our training dataset, we extract structured metadata from three RGB-D scene datasets: ScanNet~\cite{dai2017scannet}, ScanNet++~\cite{scannetpp}, and ARKitScenes~\cite{arkitscenes}. Each scene’s annotated 3D point cloud includes spatial coordinates $(x,y,z)$, RGB values, semantic labels, and instance IDs. From this, we extract geometric and semantic information essential for training. We extract the following metadata for each scene:
\begin{itemize}
    \item \textit{Scene-level attributes:} (1) Room area, computed via the convex hull of floor points; (2) Room center, defined as the centroid of all 3D points; (3) Video path, linking each scene to its corresponding RGB-D sequence.
    \item \textit{Object-level attributes:} (1) Object counts per semantic category; (2) Axis-aligned 3D bounding boxes for each instance, parameterized by centroid, spatial extents, normalized rotation axes, and min/max coordinates.
\end{itemize}
For label mapping, we handle two annotation schemes: ScanNetV2 (20 categories) and ScanNet200 (200 categories). Raw semantic labels from the point clouds are mapped to these standardized labels using the official label mapping files. We construct unified label mapping functions $f: \mathcal{L}_{\text{raw}} \rightarrow \mathcal{L}_{\text{target}}$ that convert raw label IDs to target category indices. 
% For ScanNet200, we additionally apply category remapping to merge semantically equivalent classes (e.g., merging specific furniture subtypes).

Bounding boxes are computed per instance by determining the min/max coordinates of each instance's point cluster along the three spatial axes. These boxes efficiently capture object extents and are used throughout our spatial reasoning pipeline. The extraction process is parallelized across scenes using multiprocessing to handle the large-scale datasets efficiently. Invalid or incomplete scenes (e.g., missing PLY files, corrupted annotations) are automatically filtered out during processing. Extracted metadata is stored in JSON format for efficient loading during training.

\paragraph{Load scene XY range.} 
To facilitate spatial reasoning and metric map construction, we extract the spatial boundaries of each scene from the point cloud. For each scene, we load the mesh-aligned point cloud and compute the axis-aligned bounding box that encompasses all points in the scene. Specifically, we determine the minimum and maximum coordinates along the $x$ and $y$ axes: $[x_{\min}, x_{\max}]$ and $[y_{\min}, y_{\max}]$. To ensure uniform spatial representation and simplify downstream map discretization, we normalize these bounds to form a square region. We compute the extent along each axis: $\Delta x = x_{\max} - x_{\min}$ and $\Delta y = y_{\max} - y_{\min}$. If $\Delta x > \Delta y$, we expand the $y$ bounds symmetrically by $(\Delta x - \Delta y)/2$ on each side; conversely, if $\Delta y > \Delta x$, we expand the $x$ bounds by $(\Delta y - \Delta x)/2$.

This normalization ensures that the scene occupies a square spatial region, which is essential for constructing grid-based metric cognitive maps with consistent resolution. The extracted boundaries are stored as \texttt{room\_x\_bound} and \texttt{room\_y\_bound} for each scene, representing the final square-normalized spatial extent. This preprocessing is applied uniformly across all three datasets (ScanNet, ScanNet++, and ARKitScenes) for both training and validation splits. The boundary information is saved in JSON format and later used during metric map construction to discretize continuous 3D space into a structured grid representation.

\paragraph{Make \emph{Metric-CogMap} for each scene}
To support both symbolic spatial reasoning and precise metric computation, we construct a \emph{Metric-CogMap} for each scene that encodes a dual-format representation: discrete grid coordinates for efficient relational queries and continuous metric-scale coordinates for accurate geometric analysis. Given the normalized scene boundaries $[x_{\min}, x_{\max}]$ and $[y_{\min}, y_{\max}]$, we represent each object in two parallel forms. First, for the discrete representation, we map each object's 3D centroid $(x, y, z)$ to a 2D grid cell $(g_x, g_y)$ using:
\begin{equation}
g_x = \left\lfloor \frac{x - x_{\min}}{x_{\max} - x_{\min}} \times N \right\rfloor, \quad
g_y = \left\lfloor \frac{y - y_{\min}}{y_{\max} - y_{\min}} \times N \right\rfloor
\end{equation}
, where $N$ is the grid size (default: 20), and grid indices are clipped to $[0, N-1]$. Similarly, we also compute discrete bounding boxes $([g_{x_{\min}}, g_{x_{\max}}], [g_{y_{\min}}, g_{y_{\max}}])$ by projecting the object’s axis-aligned bounding box corners to the grid.

Second, for the metric-scale representation, we store the original 3D centroid $(x, y, z)$ in meters and extract the object’s physical dimensions $(w, h, d)$ from its axis-aligned bounding box. This dual-format design allows the model to leverage discrete grids for efficient (symbolic) spatial relationship reasoning (e.g., relative direction) while enabling fine-grained metric reasoning (e.g., absolute distance) when precision is required.

For each scene, the \emph{Metric-CogMap} encodes the following structured information:
\begin{enumerate}[label=(\arabic*)]
    \item \textbf{Scene-level metadata:} The scene identifier (\texttt{scene\_id}), grid configuration (\texttt{grid\_size}, default: $20 \times 20$), and room dimensions (\texttt{room\_size}) specifying the physical extent of the scene in meters along each axis.
    \item \textbf{Object-level information:} Objects are organized by semantic category (e.g., \textit{door}, \textit{towel}, \textit{mirror}), with each category containing a list of object instances. Each object instance encodes both discrete and continuous spatial attributes:
    \begin{itemize}
        \item \texttt{instance\_id}: A unique identifier for each object instance within the scene.
        \item \texttt{grid\_position}: Discrete 2D coordinates $[g_x, g_y]$ representing the grid cell occupied by the object's plane centroid, enabling efficient spatial indexing and relationship queries.
        \item \texttt{grid\_box\_position}: Grid-aligned bounding box $[[g_{x_{\min}}, g_{x_{\max}}], [g_{y_{\min}}, g_{y_{\max}}]]$ representing the discrete spatial extent of the object in the discrete grid coordinates.
        \item \texttt{centroid}: Continuous 3D coordinates $[x, y, z]$ in meters, preserving precise metric-scale location information from the original point cloud.
        \item \texttt{bounding\_box\_size}: Physical dimensions $[w, h, d]$ in meters, computed from the axis-aligned bounding box, enabling accurate size-based reasoning.
    \end{itemize}
\end{enumerate}

This hybrid structure enables the model to bridge symbolic spatial reasoning and metric-scale computation, supporting both qualitative spatial queries and quantitative distance estimations during Cog-CoT inference. The resulting \emph{Metric-CogMap} is serialized in JSON format for seamless integration into the training pipeline.

% =================================
% Object selection for each QA pair
% =================================
\paragraph{Object selection for each QA pair}

To manage reasoning complexity, we identify the object categories referenced in each question–answer pair. This step filters the relevant entities from the scene’s \emph{Metric-CogMap}, enabling the model to focus computation on task-relevant objects rather than processing all instances in the scene. The extraction procedure relies on category-aware text parsing. For each question, we apply word–boundary–based pattern matching using the category vocabulary predefined for each dataset. Specifically, we use regex word-boundary matching ($\texttt{\textbackslash b}$\textit{category}$\texttt{\textbackslash b}$) to ensure exact category matching and avoid spurious detections (e.g., preventing “table” from matching “vegetable”).

To address cases where a shorter category name is embedded within a longer one (e.g., “box” within “tissue box”), we apply a positional filtering strategy. A category is retained only if at least one of its matched spans in the text is not entirely contained within the span of another category. This prevents us from incorrectly counting a shorter category (e.g., “box”) when it appears only as part of a longer one (e.g., “tissue box”), while still allowing both categories to be detected when they genuinely appear separately in the question. Additionally, we normalize common linguistic variations through plural-to-singular mapping (e.g., "doors" $\rightarrow$ "door") and synonym resolution (e.g., "staircase" $\rightarrow$ "stairs")improving robustness to phrasing differences.

The final output for each QA pair includes: (1) question identifier and scene name; (2) the original question text; and (3) the list of identified object categories (\texttt{found\_categories}). This structured representation enables the model to retrieve only the relevant object instances from the scene's \emph{Metric-CogMap} during inference, significantly reducing computational overhead and improving precision in spatially grounded reasoning.

% =================================
% Create Metric-CogMap for each QA pair.
% =================================
\paragraph{Create \emph{Metric-CogMap} for each QA pair.}

Once the relevant object categories are identified for a given question, we construct a question-specific \emph{Metric-CogMap} by extracting only the corresponding object instances from the full scene-level map. This targeted subset serves as the spatial context for the question, reducing noise and focusing the model’s attention on task-relevant elements. The resulting map is embedded into the input prompt alongside task-specific reasoning instructions, enabling efficient and interpretable spatial reasoning.

\noindent \textbf{(1) Question-specific map construction.} For each QA pair, we retrieve all instances of the identified categories from the scene's complete \emph{Metric-CogMap} to create a compact, question-specific JSON representation. This tailored map includes four components:

\begin{itemize}
    \item \textbf{Cognitive map:} Discrete grid positions $[g_x, g_y]$ for each object instance, e.g., \texttt{\{"trash bin": [[5, 13]], "nightstand": [[14, 11]]\}}.
    \item \textbf{Cognitive box map:} Grid-aligned bounding boxes $[[g_{x_{\min}}, g_{x_{\max}}], [g_{y_{\min}}, g_{y_{\max}}]]$, e.g., \texttt{\{"trash bin": [[[4, 5], [12, 13]]]\}}.
    \item \textbf{Box centroid:} Continuous metric 3D coordinates $[x, y, z]$ in meters, e.g., \texttt{\{"trash bin": [[-1.5, 0.58, 0.12]]\}}.
    \item \textbf{Box size:} Physical dimensions $[w, h, d]$ in meters, e.g., \texttt{\{"trash bin": [[0.36, 0.30, 0.37]]\}}.
\end{itemize}

\noindent \textbf{(2) Task-specific reasoning Instructions.} The question-specific \emph{Metric-CogMap} is serialized in JSON format and appended to the original question text. To enable explicit spatial reasoning, we additionally inject task-specific computational instructions that explicitly instruct the model on the required mathematical operations. Below are representative examples:

\begin{itemize}
    \item \textit{Relative direction:} ``\textit{Task: Determine relative direction (front-left/front-right/back-left/back-right) using vector analysis based on cognitive map. Format: Origin = [x,y], Facing = [x,y], Target = [x,y] $\rightarrow$ f = Facing-Origin $\rightarrow$ t = Target-Origin $\rightarrow$ Dot(f,t) for front/back $\rightarrow$ Cross(f,t) for left/right. Output: Mathematical calculations with explicit coordinate subtraction, dot/cross product computation, and directional conclusion.}''
    
    \item \textit{Relative distance:} ``\textit{Task: Compare distances between multiple candidates and a target to find the closest one based on cognitive map and cognitive box map. Format: Target = [x,y], Candidates = [obj1, obj2, ...] $\rightarrow$ Distance(obj1,target) = calculation $\rightarrow$ Distance(obj2,target) = calculation $\rightarrow$ Compare all distances $\rightarrow$ Closest object = result. Output: Show distance calculations for each candidate with explicit mathematical steps and final comparison.}''
    
    \item \textit{Absolute distance:} ``\textit{Task: Calculate absolute distance between two specific objects based on box centroid and box size. Format: Object1: Centroid c1 = [x,y,z], Size = [w,h,d], Half size s1 $\rightarrow$ Object2: Centroid c2, Size, Half size s2 $\rightarrow$ Centroid difference $\rightarrow$ Half size sum $\rightarrow$ Rough distance calculation. Output: Step-by-step box data extraction, centroid calculations, and rough distance approximation with explicit numerical values.}''

    \item \textit{Room size:} ``\textit{Task: Estimate room area using the number of valid cells calculation. Format: Number valid cells $\rightarrow$ Multiply by each cell area (0.36 m²).  Answer with format like \texttt{\textless answer\textgreater final\_room\_size\textless /answer\textgreater}. Output: Explicit calculations with multiplication, rough room size approximation.}''

\end{itemize}

This structured representation provides the model with both the spatial data and explicit computational guidance needed for grounded spatial reasoning.

% \section{Complete QA, \emph{Metric-CogMap}, CogCoT}
\section{\emph{Cog-CoT} construction and Complete QA.}
\label{Supp:full_cogmap_cogcot}

% =================================
% Create Cog-CoT for each QA pair.
% =================================
\paragraph{Generating \emph{Cog-CoT} for each QA pair.} 
To enable the model to perform explicit spatial reasoning over the \emph{Metric-CogMap}, we augment each training QA pair with programmatically generated chain-of-thought (Cog-CoT) supervision. This supervision encodes step-by-step spatial computations tailored to the reasoning requirements of each question type.

\paragraph{Programmatic CoT generation.}
For each question category, we design dedicated algorithms that execute the relevant geometric operations on the \emph{Metric-CogMap} to generate verifiable intermediate reasoning steps.

\begin{figure*}[htpb]
	\centering
	\includegraphics[width=1.0 \linewidth]{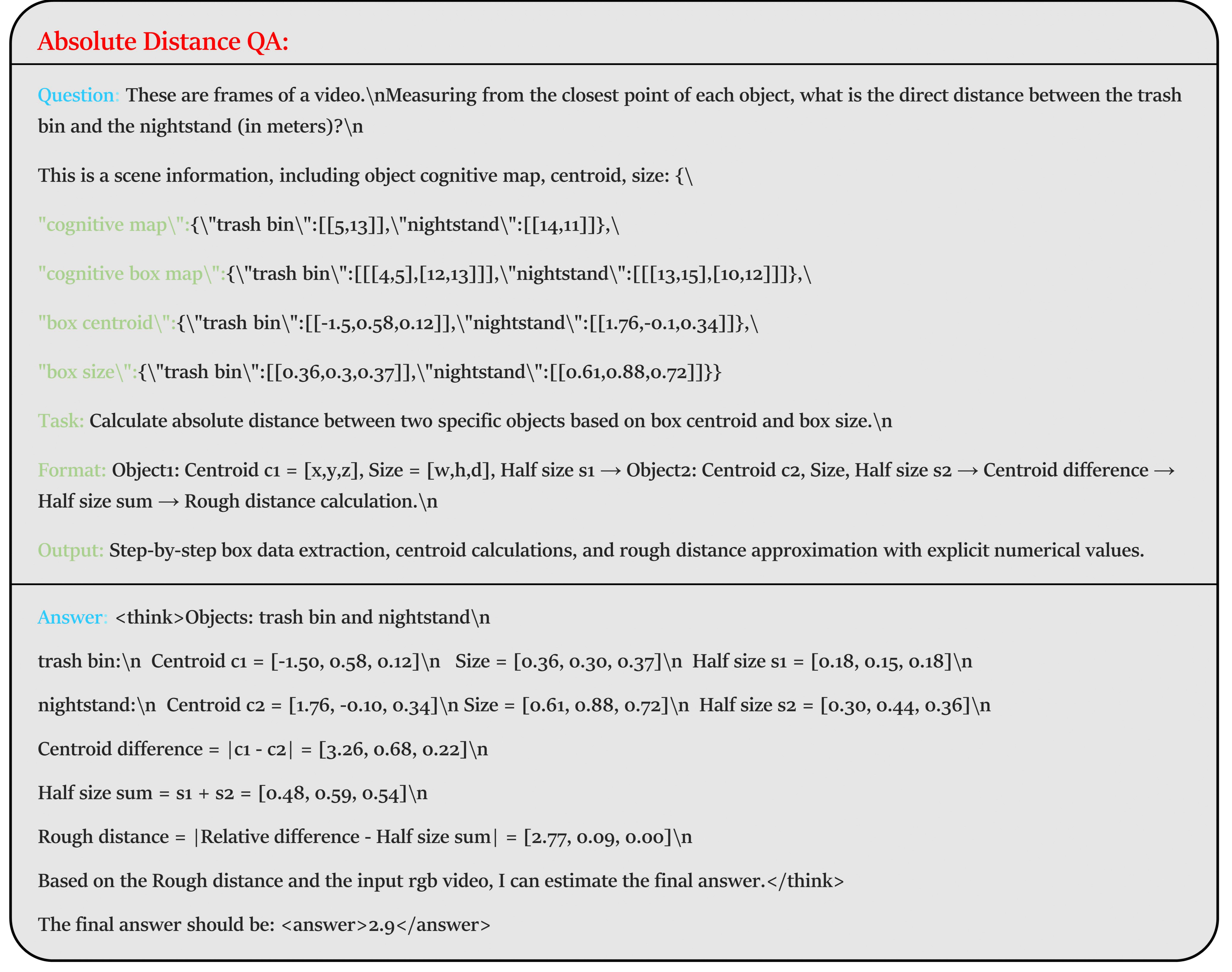} \\
    \caption{\textbf{Examples of \emph{Metric-CogMap} and \emph{CogCoT} on absolute\_distance}}
    \label{fig:supp_qa_abs_dist}
\end{figure*}

\begin{figure*}[htpb]
	\centering
    \includegraphics[width=1.0 \linewidth]{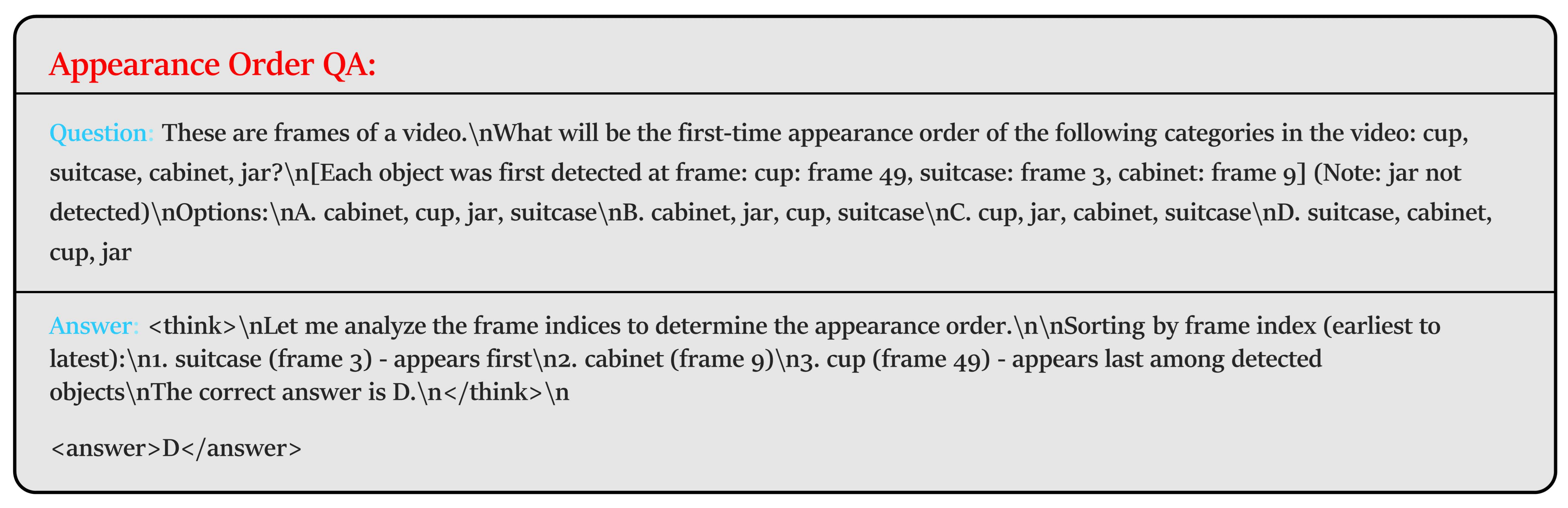} \\
    \caption{\textbf{Examples of \emph{Metric-CogMap} and \emph{CogCoT} on appearance\_order}}
    \label{fig:supp_qa_app_order}
\end{figure*}

\begin{enumerate}[label=(\alph*)]
    \item \textbf{Relative direction CoT:} The algorithm extracts object coordinates from the cognitive map and computes the facing vector $\mathbf{f} = \text{Facing} - \text{Origin}$ and target vector $\mathbf{t} = \text{Target} - \text{Origin}$ with explicit coordinate subtraction. It then calculates the dot product $\mathbf{f} \cdot \mathbf{t} = f_x \cdot t_x + f_y \cdot t_y$ to determine whether the target is in front ($> 0$) or behind ($< 0$), and the 2D cross product $f_x t_y - f_y t_x$ to determine whether the target lies to the left ($> 0$) or right ($< 0$), outputting all intermediate numerical values.
    
    The complete question, \emph{Metric-Cogmap}, \emph{Cog-CoT}, and answer for \emph{relative direction} is illustrated in Fig.~\ref{fig:supp_qa_rel_dir}.
    
\begin{figure*}[htpb]
	\centering
    \includegraphics[width=1.0 \linewidth]{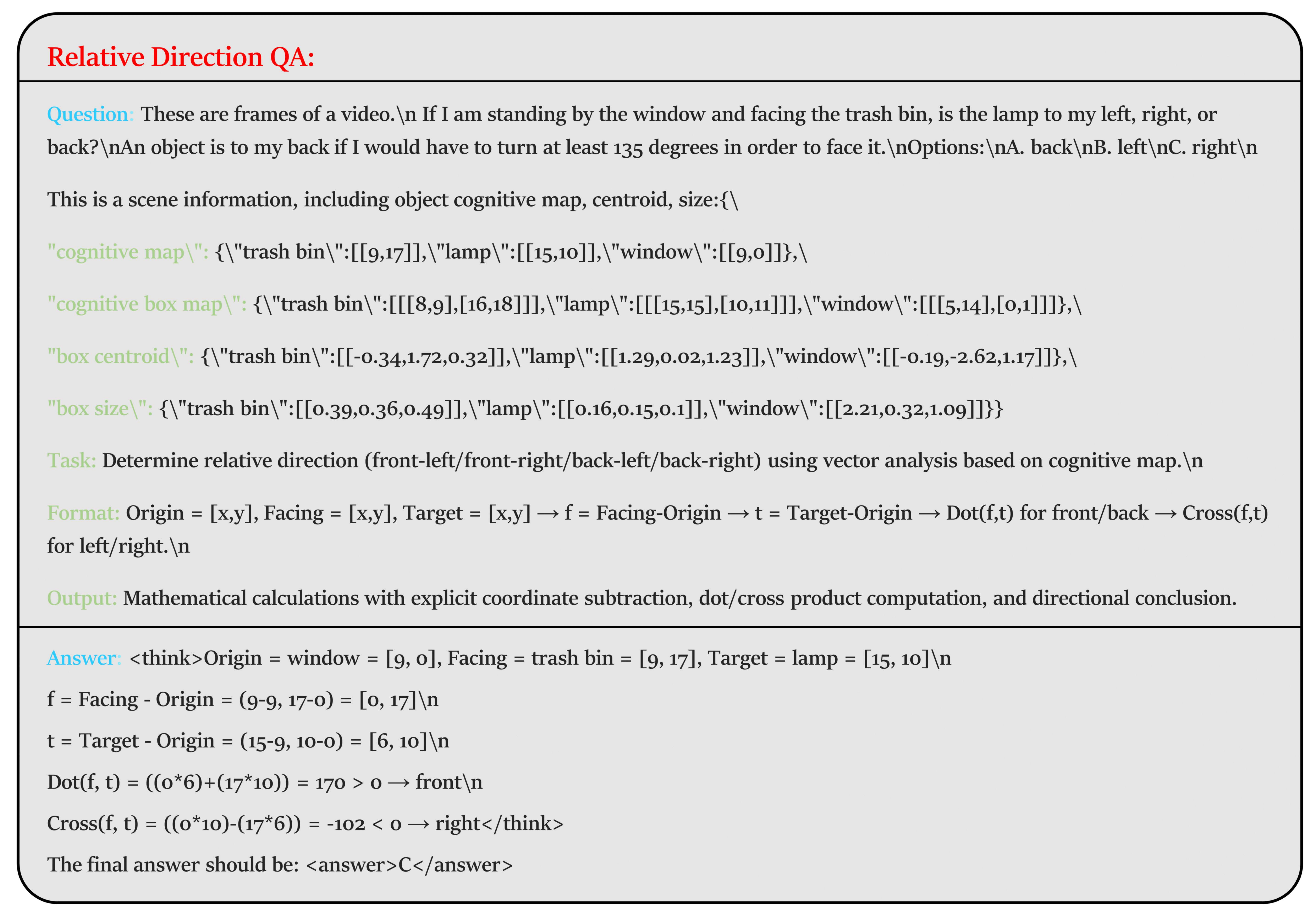} \\
    \caption{\textbf{Examples of \emph{Metric-CogMap} and \emph{CogCoT} on relative\_direction}}
    \label{fig:supp_qa_rel_dir}
\end{figure*}

\begin{figure*}[htpb]
	\centering
    \includegraphics[width=1.0 \linewidth]{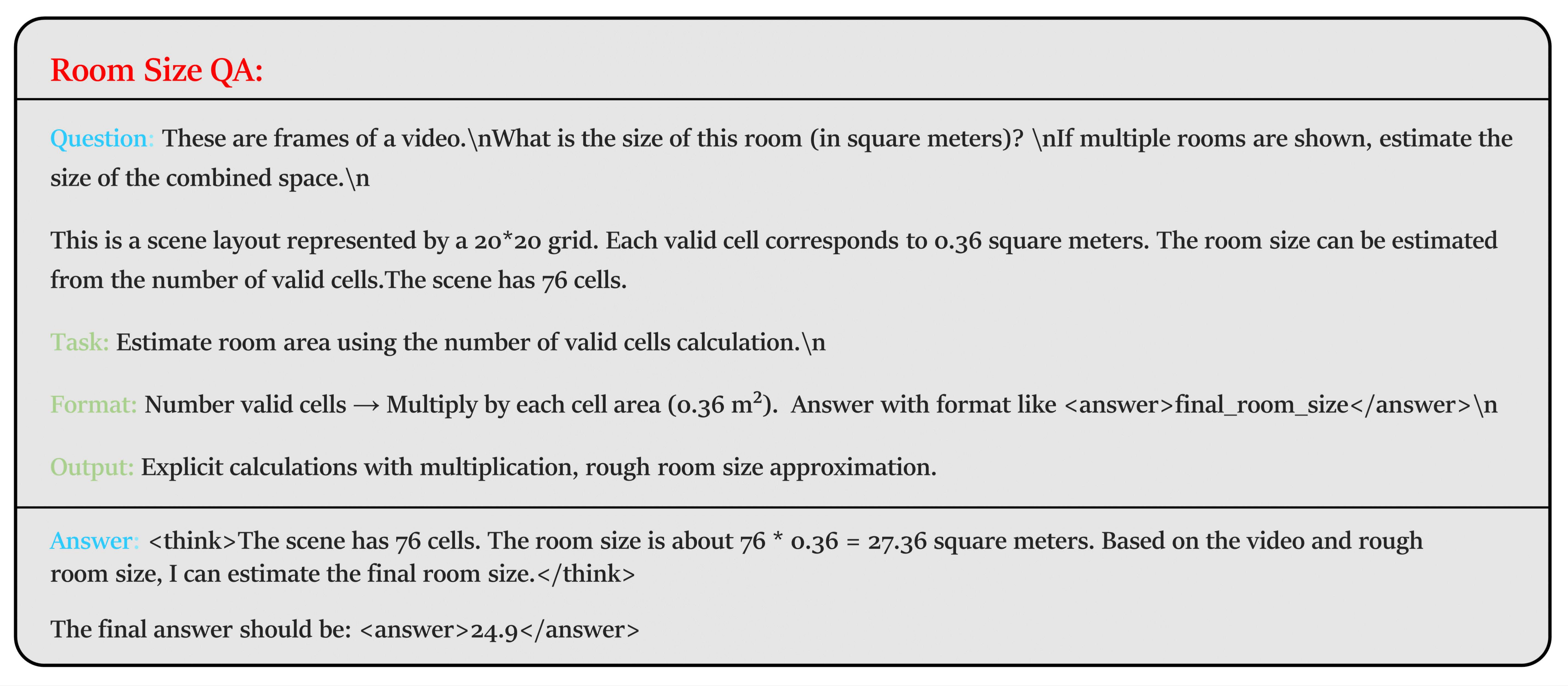} \\
    \caption{\textbf{Examples of \emph{Metric-CogMap} and \emph{CogCoT} on room\_size}}
    \label{fig:supp_qa_room_size}
\end{figure*}
    
    \item \textbf{Room size CoT:} The algorithm operates on the $20 \times 20$ occupancy grid, where each cell corresponds to an area of $0.36 m^{2}$. It counts the number of valid (occupied) cells, multiplies by the per-cell area to compute the total room size, and explicitly shows the calculation steps (e.g., ``Total cells = 76, total area = 76 $\times$ 0.36 = $27.36 m^{2}$'').
    
    The complete question, \emph{Metric-Cogmap}, \emph{Cog-CoT}, and answer for \emph{room size} is illustrated in Fig.~\ref{fig:supp_qa_room_size}.

    \item \textbf{Absolute distance CoT:} To estimate the distance between two specified objects, the algorithm retrieves their centroids $c_1 = [x_1, y_1, z_1]$ and $c_2 = [x_2, y_2, z_2]$ from the box centroid data, and their sizes from the box size data. It computes half-sizes $s_1 = \text{Size}_1 / 2$ and $s_2 = \text{Size}_2 / 2$ component-wise, calculates the absolute centroid difference $\Delta = |c_1 - c_2|$, sums the half-sizes $s = s_1 + s_2$, and estimates the approximate object-to-object distance as $\text{dist} = \max(\Delta - s, 0)$ component-wise. All arithmetic operations are explicitly shown with numerical values (e.g., ``Centroid difference = $\|[-1.50, 0.58, 0.12] - [1.76, -0.10, 0.34]\| = [3.26, 0.68, 0.22]$'').
    
    The complete question, \emph{Metric-Cogmap}, \emph{Cog-CoT}, and answer for \emph{absolute distance} is illustrated in Fig.~\ref{fig:supp_qa_abs_dist}.
    
    \item \textbf{Appearance order CoT:} To infer appearance order, we leverage the sparse 3D point clouds stored in the \emph{Metric-CogMap}, which are quantized per object instance using the predicted camera poses provided by $\pi^{3}$. Since precise geometry is not strictly required for this task, we use the precomputed quantized point clouds and project them onto each of the 64 subsampled frames used as input to the VLM. For each object instance, we project its 3D points into the image plane of each frame and record the first frame index (ranging from 0 to 63) where the object becomes visibly present. To avoid false positives caused by occlusions (e.g., when a wall or another object blocks the target instance), we apply a visibility filter: projected points are discarded if there exists a reconstructed (unlabeled) point in front of them—closer to the camera by more than a threshold of 0.2 meters (20 centimeters in metric scale). This filtering ensures that only unoccluded instances are considered when identifying appearance frames.
    
    For temporal ordering questions, the algorithm receives frame detection indices for each object (e.g., ``door: frame 30, printer: frame 249, mouse: frame 253''), sorts the objects in ascending order by frame index, and determines their relative order (e.g., which appeared first or last).
    In cases of missing detections (e.g., ``kettle not detected''), the reasoning chain explicitly includes a note about the absence, allowing the model to learn inference under incomplete observations. The reasoning concludes by matching the derived order against multiple-choice options. 
    % he resulting frame indices are then used to build a \emph{Cog-CoT} chain for appearance order reasoning, as illustrated in Fig.~\ref{fig:supp_qa_app_order}.
    
    The complete question, \emph{Metric-Cogmap}, \emph{Cog-CoT}, and answer for \emph{appearance order} is illustrated in Fig.~\ref{fig:supp_qa_app_order}.

    \item \textbf{Relative distance CoT:} For queries that ask which object is closest among several candidates, the algorithm computes axis-aligned bounding box (AABB) distances. For each candidate category, it retrieves all object instances from the \emph{Metric-CogMap}. Then, for each instance–target pair, it calculates per-axis separation: if the boxes do not overlap on axis $x$, then $d_x = \max(x_{\text{cand}}^{\min} - x_{\text{tgt}}^{\max}, x_{\text{tgt}}^{\min} - x_{\text{cand}}^{\max})$; otherwise, $d_x = 0$. The squared AABB distance $d^2 = d_x^2 + d_y^2$ is computed for each instance, showing detailed calculations (e.g., ``dx = 15 - 14 = 1, dy = 0 (y-axis overlap), AABB distance = $1^2 + 0^2 = 1.0$''). The minimum distance is selected per category, and all candidates are compared to identify the closest object.
    
    The complete question, \emph{Metric-Cogmap}, \emph{Cog-CoT}, and answer for \emph{relative distance} is illustrated in Fig.~\ref{fig:supp_qa_rel_dist}.
\end{enumerate}

\paragraph{Structured response format.} Each training sample's ground-truth response is structured as: 

\texttt{<think>step-by-step reasoning with numerical computations</think> The final answer should be: <answer>answer</answer>}. 

This format enables the model to learn both the spatial reasoning process and the final answer prediction jointly. By providing explicit computational demonstrations through Cog-CoT, the model learns to decompose complex spatial questions into verifiable symbolic operations over the \emph{Metric-CogMap} representation, effectively bridging natural language understanding with geometric reasoning.

\begin{figure*}[htpb]
	\centering
    \includegraphics[width=1.0 \linewidth]{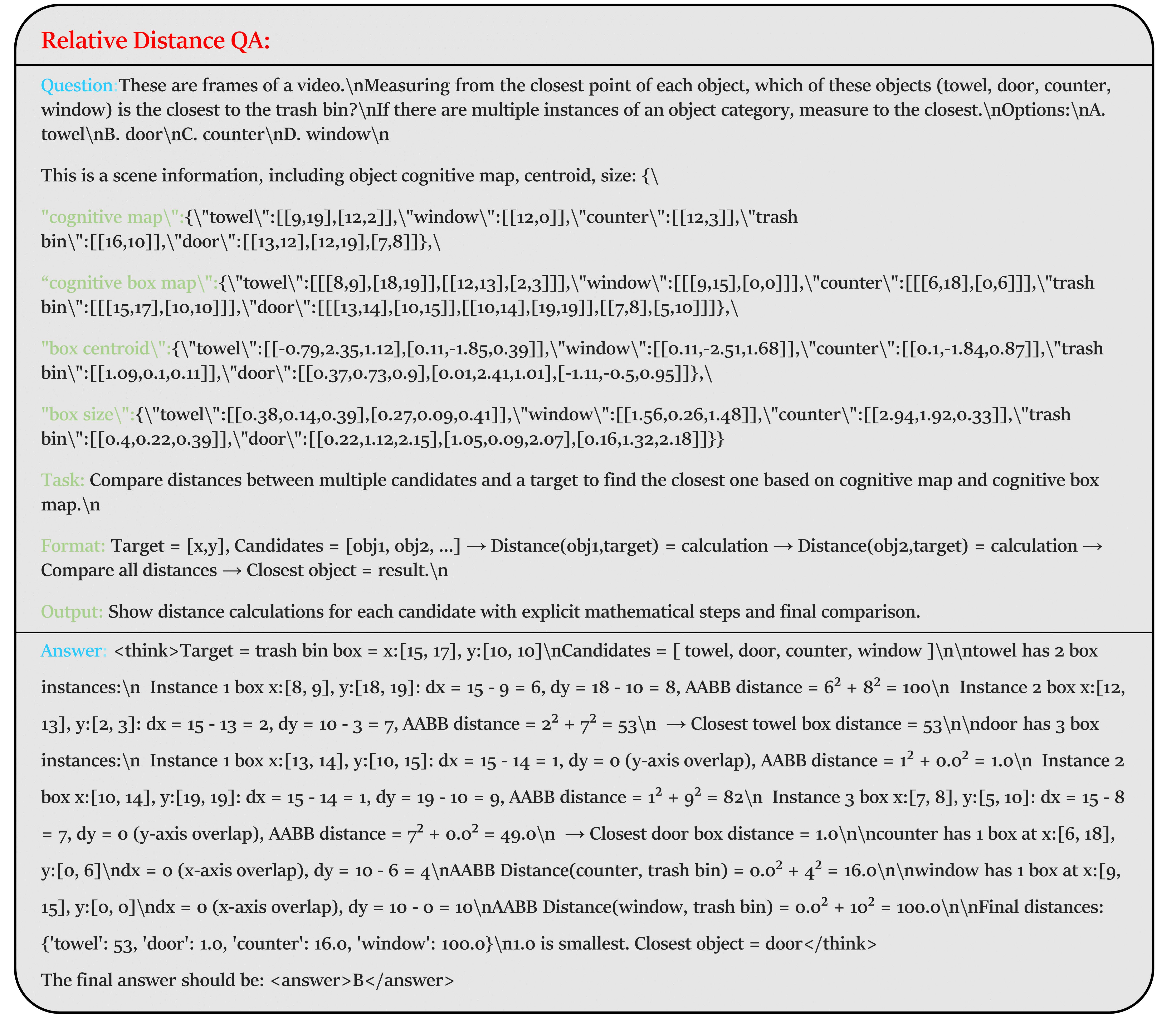} \\    
    \caption{\textbf{Examples of \emph{Metric-CogMap} and \emph{CogCoT} on relative\_distance}}
    \label{fig:supp_qa_rel_dist}
\end{figure*}

% \noindent \textbf{(a) Obj. Count}.

% \noindent \textbf{(b) Abs. Dist}.

% \noindent \textbf{(c) Obj. Size}.

% \noindent \textbf{(d) Room Size}.

% \noindent \textbf{(e) Rel. Dist}.

% \noindent \textbf{(f) Route Plan}.

% \noindent \textbf{(g) Apprance. Order}. To infer appearance order, we leverage the sparse 3D point clouds stored in the \emph{Metric-CogMap}, which are quantized per object instance using the predicted camera poses provided by $\pi^{3}$. Since precise geometry is not strictly required for this task, we use the precomputed quantized point clouds and project them onto each of the 64 subsampled frames used as input to the VLM. For each object instance, we project its 3D points into the image plane of each frame and record the first frame index (ranging from 0 to 63) where the object becomes visibly present. To avoid false positives caused by occlusions (e.g., when a wall or another object blocks the target instance), we apply a visibility filter: projected points are discarded if there exists a reconstructed (unlabeled) point in front of them—closer to the camera by more than a threshold of 0.2 meters (20 centimeters in metric scale). Using this filtered visibility analysis, we identify the earliest visible frame for each queried object mentioned in the text. The resulting frame indices are then used to build a \emph{Cog-CoT} chain for appearance order reasoning, as illustrated in Fig.~\ref{fig:supp_qa_app_order}{(g)}.

% \section{Details on \emph{Metric-CogMap} in evaluation}

\section{Zero-shot qualitative demonstrations}
\label{Supp:add_quantitative}
We conducted a small-scale zero-shot evaluation to assess whether augmenting a generic vision-language model with our proposed \emph{Metric-CogMap} and \emph{Cog-CoT} prompts improves spatial reasoning, even without task-specific fine-tuning. To this end, we tested a few randomly selected examples using publicly available commercial models, following the same input protocol as VSI-Bench. Specifically, we used a zero-shot setting in which only 10 subsampled video frames (the maximum allowed for upload) and the test query were provided. 

\begin{figure*}[htpb]
	\centering
	\begin{subfigure}[b]{1.0\linewidth}
		\includegraphics[width=\linewidth]{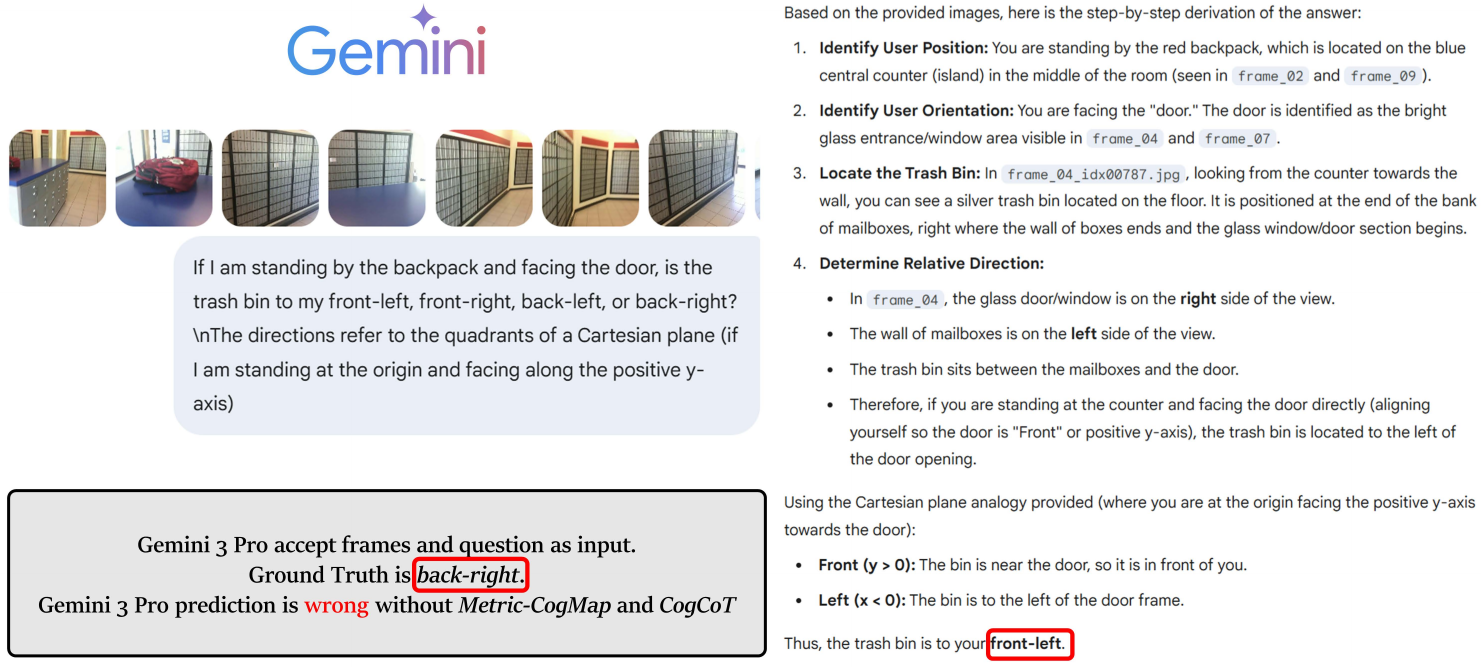}
		\caption{Baseline zero-shot result from Gemini-3-Pro using only image and text query inputs. The model fails to interpret the spatial relationship.}
	\end{subfigure}
	
	\vspace{0.5em}
	%\hrulefill
	\vspace{0.5em}
	
	\begin{subfigure}[b]{1.0\linewidth}
		\includegraphics[width=\linewidth]{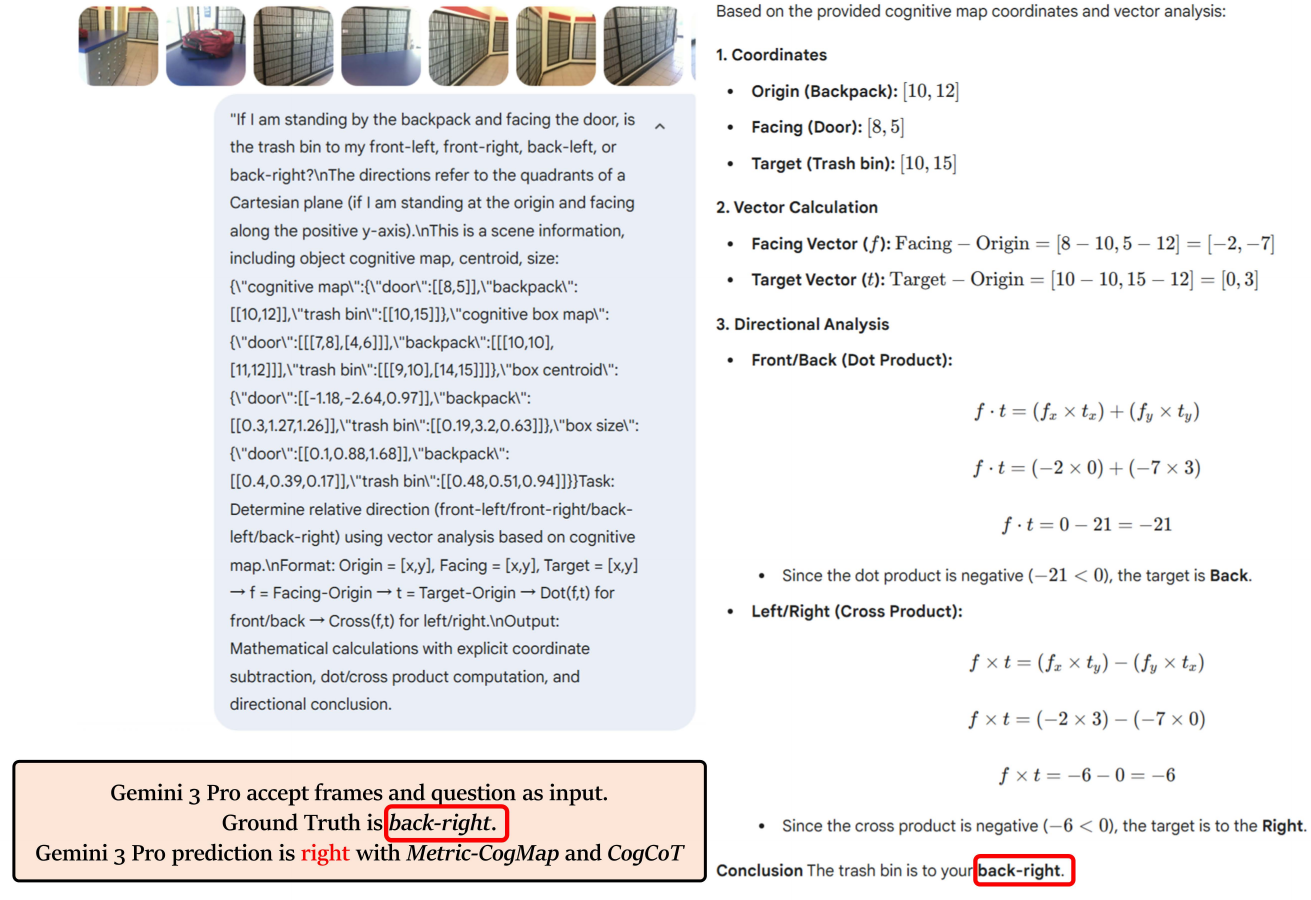}
		\caption{Our zero-shot result using the same model augmented with \emph{Metric-CogMap} and \emph{Cog-CoT}. The model correctly reasons and answers the spatial query.}
	\end{subfigure}
	
    \caption{\textbf{Qualitative comparison of zero-shot spatial direction reasoning.} This example demonstrates the effect of embedding \emph{Metric-CogMap} and \emph{Cog-CoT} into a frozen Gemini-3-Pro model. While the baseline fails to resolve the \texttt{relative\_direction} query using only image and text input, our prompt-enhanced approach enables successful geometric reasoning and accurate answer prediction.}

	\label{fig:zeroshot}
\end{figure*}

\begin{figure*}[htpb]
	\centering
	\begin{subfigure}[b]{1.0\linewidth}
		\includegraphics[width=\linewidth]{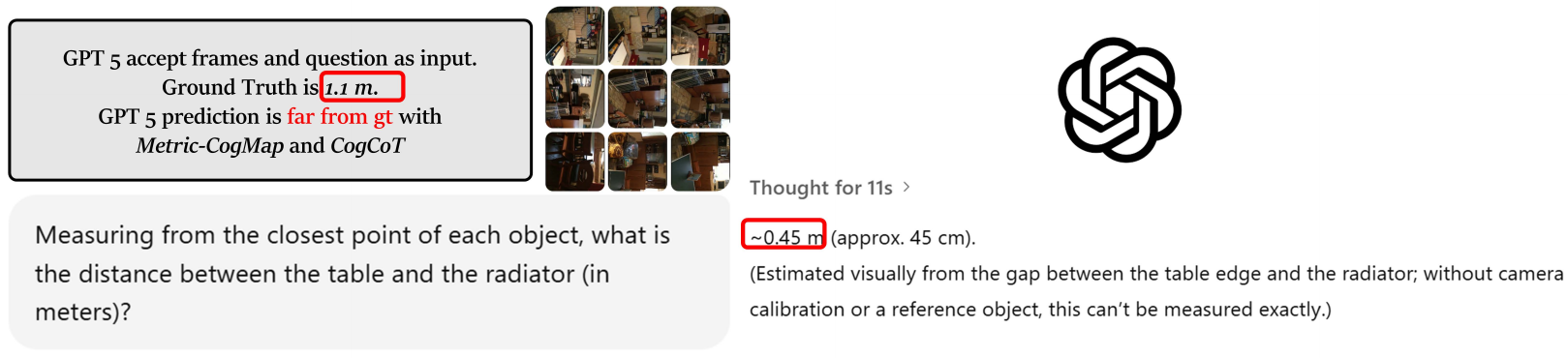}
		\caption{Baseline zero-shot result from ChatGPT-5-Thinking using only image and text query inputs. The model fails to measure metric distance.}
	\end{subfigure}
	
	\vspace{0.5em}
	%\hrulefill
	\vspace{0.5em}
	
	\begin{subfigure}[b]{1.0\linewidth}
		\includegraphics[width=\linewidth]{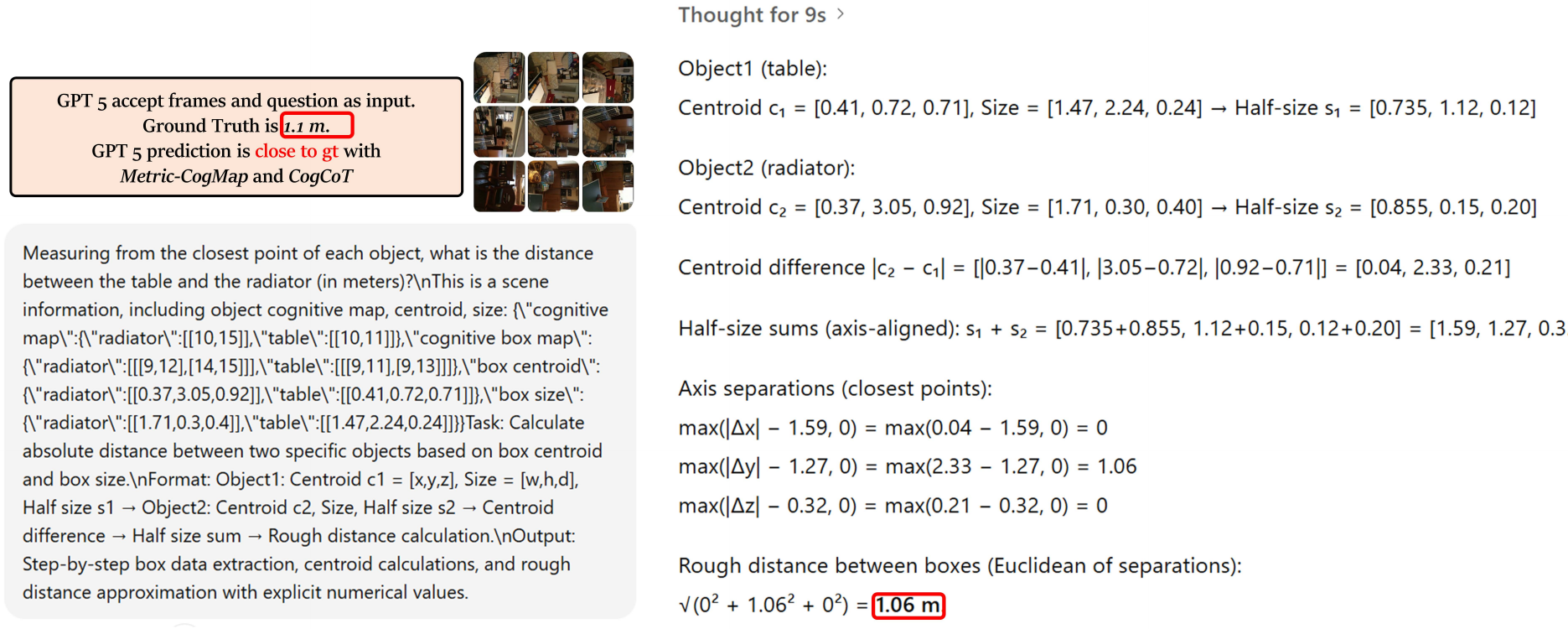}
		\caption{Our zero-shot result using the same model augmented with \emph{Metric-CogMap} and \emph{Cog-CoT}. The model correctly calculates the distance and answers the spatial query.}
	\end{subfigure}
	
    \caption{\textbf{Qualitative comparison of zero-shot absolute distance estimation.} This example shows the impact of embedding \emph{Metric-CogMap} and \emph{Cog-CoT} into a frozen GPT-5-Thinking model. The baseline model fails to compute correct spatial distance in the \texttt{absolute\_distance} task, while our zero-shot-enhanced prompt enables metric scale reasoning and correct inference.}

	\label{fig:zeroshot_gpt}
\end{figure*}

Figs.~\ref{fig:zeroshot} and \ref{fig:zeroshot_gpt} illustrate comparisons between two setups for Gemini-3-Pro and GPT-5-Thinking, respectively: (a) using only images and the question as input, and (b) the same models augmented with \emph{Metric-CogMap} and \emph{Cog-CoT} prompts. In Fig.~\ref{fig:zeroshot}, the Gemini-3-Pro baseline fails to answer a relative direction question, while the ours successfully infers the correct spatial relation. Similarly, in Fig.~\ref{fig:zeroshot_gpt}, the GPT-5-Thinking baseline struggles with distance estimation, whereas our zero-shot-enhanced prompt enables it to reason accurately.

Due to the limitations of commercial APIs, we could not perform full VSI-Bench evaluation. However, these preliminary examples demonstrate that our Map2Thought framework improves spatial understanding in zero-shot scenarios—even when applied to state-of-the-art proprietary models—highlighting the value of explicit spatial reasoning guidance for models that otherwise lack this capability.

% \section{Extended benchmarks}
% \label{Supp:ext_benchmarks}

% {
%     \small
%     \bibliographystyle{ieeenat_fullname}
%     \bibliography{main}
% }


\end{document}